\documentclass[preprint,11pt]{elsarticle}

\usepackage{hyperref}
\usepackage{breakurl}
\usepackage{amssymb}
\usepackage{tikz}
\usetikzlibrary{positioning,decorations.pathreplacing,shapes}
\usepackage{dirtytalk}
\usepackage{amsmath}
\usepackage{amsthm}
\usepackage{algorithm}
\usepackage{algpseudocode}
\usepackage{bbm}
\usepackage{newfloat}
\usepackage{listings}
\usepackage{amsfonts,amsmath}
\usepackage{bm}
\usepackage{hyperref}
\usepackage{cleveref}
\usepackage{thmtools,thm-restate}
\usepackage{caption}
\usepackage{subcaption}
\usetikzlibrary{calc}
\newtheorem{definition}{Definition}
\newtheorem{example}{Example}
\newtheorem{theorem}{Theorem}
\newtheorem{remark}{Remark}
\newtheorem{proposition}{Proposition}
\newtheorem{lemma}{Lemma}
\newtheorem*{proposition*}{Proposition}

\usepackage{times}
\usepackage{url}
\usepackage{xcolor}
\usepackage{yhmath}
\usepackage{cases}
\usepackage{bm}
\usepackage{bbm}
\usepackage{dirtytalk}
\usepackage{mathtools} % amsmath with fixes and additions
\usepackage{booktabs} % commands to create good-looking tables
\usepackage{amsmath}
\usepackage{amsthm}
\usepackage{algorithm}
\usepackage{algpseudocode}
\usepackage{bbm}
\usepackage{newfloat}
\usepackage{listings}
\usepackage{amsfonts,amsmath}
\usepackage{bm}
\usepackage{hyperref}
\usepackage{cleveref}
\usepackage{thm-restate}

\def\C{{Connected(R)}}

\def\bx{\bm{x}}

\def\fomc{\mbox{\sc fomc}}

\def\wfomc{\mbox{\sc wfomc}}
\def\w{\mbox{\sc w}}

\def\R{\mathbb{R}}

\def\bk{{\bm{k}}}
\def\bp{{\bm{p}}}

\def\bs{{\bm{s}}}

\RequirePackage{bm}
\RequirePackage{xcolor}

\def\bx{\bm{x}}

\def\fomc{\mbox{\sc fomc}}

\def\wfomc{\mbox{\sc wfomc}}

\def\R{\mathbb{R}}

\def\bk{{\bm{k}}}

\newcommand{\red}[1]{\textcolor{red}{#1}}

\def\x{\bm x}

\usepackage{etoolbox,environ}

\newtoggle{comm}
\toggletrue{comm}
\NewEnviron{comment}
  {\iftoggle{comm}{\BODY}{}}

\usepackage{comment}
\includecomment{arxiv}
\includecomment{AIJ}
\excludecomment{AIJ}

 \journal{the Journal of Artificial Intelligence}

\begin{document}

\begin{frontmatter}



\title{Lifted Inference beyond First-Order Logic}


\author[inst1]{Sagar Malhotra}

\affiliation[inst1]{organization={TU Wien},%Department and Organization
            city={Vienna},
            country={Austria}}

\author[inst2,inst3]{Davide Bizzaro}
\author[inst2]{Luciano Serafini}

\affiliation[inst2]{organization={Fondazione Bruno Kessler},%Department and Organization
            addressline={Trento}, 
            country={Italy}}

\affiliation[inst3]{organization={University of Padova},%Department and Organization
            addressline={Padova}, 
            country={Italy}}

\begin{abstract}

  Weighted First Order Model Counting (WFOMC) is fundamental to probabilistic inference in statistical relational learning models. As WFOMC is known to be intractable in general ($\#$P-complete), logical fragments that admit polynomial time WFOMC are of significant interest. Such fragments are called \emph{domain liftable}. Recent works have shown that the two-variable fragment of first order logic extended with counting quantifiers (C$^2$) is domain-liftable. However, many properties of real-world data, like \emph{acyclicity} in citation networks and \emph{connectivity} in social networks, cannot be modeled in C$^2$, or first order logic in general. In this work, we expand the domain liftability of C$^2$ with  multiple such properties. We show that any C$^2$ sentence remains domain liftable when one of its relations is restricted to represent a directed acyclic graph, a connected graph, a tree (resp. a directed tree) or a forest (resp. a directed forest). All our results rely on a novel and general methodology of \emph{counting by splitting}. Besides their application to probabilistic inference, our results provide a general framework for counting combinatorial structures. We expand a vast array of previous results in discrete mathematics literature on directed acyclic graphs, phylogenetic networks, etc. 
  \end{abstract}

\begin{highlights}
\item We expand domain-liftability of C$^2$, with constraints not expressible in FOL.

\item Our constraints include: acyclicity, connectivity and  forests.
\item We introduce a novel principle of \say{counting by splitting} for extending WFOMC.
\item We provide an extensive set of combinatorial and relational AI applications.

\end{highlights}

\begin{keyword}
Lifted Inference \sep Enumerative Combinatorics \sep Weighted Model Counting \sep First Order Logic \sep Directed Acyclic Graphs \sep Connected Graphs \sep Counting Quantifiers \sep Statistical Relational Learning 
% MSC codes here, in the form: \MSC code \sep code
% or \MSC[2008] code \sep code (2000 is the default)
\end{keyword}

\end{frontmatter}

%%%%%%%%%%%%%%%%%%%%%%%%%%%% INTRODUCTION %%%%%%%%%%%%%%%%%%%%%%%%%%%%%%%%%%%%%%%%%%%%%%%
\section{Introduction}

Statistical Relational Learning (SRL) \cite{SRL_LISA, SRL_LUC} is concerned with modeling, learning and inferring over relational data. Probabilistic inference and learning in SRL models like  Markov Logic Networks (MLN) \cite{richardson2006markov} and Probabilistic Logic Programs (PLP) \cite{PLP} can be reduced to instances of \emph{Weighted First Order Model Counting} (WFOMC) \cite{broeck2013, PTP}. WFOMC is the task of computing the weighted sum of the models of a First Order Logic (FOL) sentence $\Phi$ over a finite domain of size $n$. Formally, 
\begin{equation}
    \label{eq: WFOMC}
    \wfomc(\Phi,\w,n) = \sum_{\omega\models\Phi}\w(\omega)
\end{equation}
where $\w$ is a \emph{weight function} that associates a real number to each interpretation $\omega$. Fragments of FOL that admit polynomial time WFOMC w.r.t the domain cardinality $n$ are known as \emph{domain-liftable} \cite{domain_lifted_defintion} or equivalently are said to admit \emph{lifted-inference}.  Recent works have also explored WFOMC as a tool for dealing with combinatorics, providing closed form formulae for counting combinatorial structures \cite{Symmetric_Weighted, kuusisto2018weighted,AIxIA, AAAI_Sagar} and identifying novel combinatorial sequences \cite{Automatic_conjecturing, discovering_sequences}. This generality of WFOMC applications has led to significant interest in FOL fragments that are domain liftable \cite{LIFTED_PI_KB_COMPLETION,DOMAIN_RECURSION,kazemi2016new}. 

A series of results \cite{Symmetric_Weighted,broeck2013} have led to a rather clear picture of domain-liftability in the two-variable fragment of FOL. Especially, these results have  shown that any FOL formula in the two-variable fragment is domain-liftable \cite{Symmetric_Weighted}. These positive results are also accompanied by intractability results \cite{Symmetric_Weighted,Jaeger_lifted}, showing that there is an FOL formula in the three-variable fragment whose (W)FOMC can not be computed in polynomial time. Hence, a significant effort has been made towards expanding the domain-liftability of the two-variable fragment of FOL with additional constraints \cite{kuusisto2018weighted,kuzelka2020weighted, AAAI_Sagar,Linear_Order_Axiom, Tree}. However, many features of real-world data are still not captured in these domain-liftable fragments. Among such constraints are \emph{acyclicity} and \emph{connectivity}. Both of them are ubiquitous features of real-world data \cite{DAG_Citation, Ugander2011TheAO}, and have been investigated extensively in combinatorics literature \cite{counting_DAG_Source_Sink, robinson1973counting, Counting_Forests}. Therefore, extending domain-liftable fragments of FOL with these constraints is of both theoretical and practical interest. However, both acyclicity and connectivity cannot be expressed in FOL \cite{immerman2012descriptive}. Formally, this means that there is no FOL sentence with fixed number of variables or quantifiers, that can express that a graph is acyclic or connected for arbitrary number of nodes.

In this work, we introduce a general principle of \emph{counting by splitting} for relational structures (Section \ref{sec:main}). 
The principle is based on the simple observation that any interpretation preserves the satisfaction of universally quantified FOL sentences when projected onto a subset of the domain (Proposition~\ref{prop: formula on subset}). 
Conversely, interpretations of a given universally quantified FO$^2$ sentence $\Phi$ over disjoint sets of domain constants can be expanded (in different ways) to a new interpretation on the union of the sets, such that $\Phi$ and some additional desired properties are satisfied (as evidenced in Lemma~\ref{lem: count_split} and ~\ref{lem: count_split_wmc}).
Using counting by splitting with existing combinatorial approaches, we expand the domain-liftability of the universally quantified formulas in the two-variable fragment of FOL (FO$^2$) with directed acyclicity (Section~\ref{sec: DAG}) and connectivity constraints (Section~\ref{sec: connected}). Formally, we show that  a purely universally quantified FO$^2$ sentence, with  a distinguished predicate restricted to represent a directed acyclic graph or a connected graph, is domain-liftable. 
We expand these results to  full FO$^2$, extended with \emph{cardinality constraints}, and \emph{counting quantifiers} (C$^2$), using existing techniques in the literature \cite{broeck2013, kuzelka2020weighted}. 
The combination of directed acyclicity, connectivity, and counting quantifiers, allow easy extension of domain-liftability of C$^2$, with tree, directed tree and directed forest constraints (Section \ref{sec: forest}).  
Finally, forest constraints are obtained by using tree constraints, and counting by splitting.

Besides its applications to SRL, our work provides a formal language in which any expressed constraint can be counted tractably, offering a valuable tool for solving problems in enumerative combinatorics \cite{Herbert_Wilf}.
Most algorithms in this field are devised on a case-by-case basis, almost independently for each problem. In contrast, WFOMC with graph-theoretical constraints offers a general framework: with it, the main task reduces to formulating a logical expression for each problem --- something that is significantly easier than devising individual counting algorithms.
We illustrate this point with multiple examples of problems investigated in the combinatorics literature that can be easily expressed as the WFOMC of C$^2$ formulas with the presented constraints. 
To the best of our knowledge, we show for the first time that the number of \emph{binary phylogenetic networks} on $n$ nodes can be counted in polynomial time w.r.t the number of nodes (see Example~\ref{ex:phylogenetic}).

Before presenting our main results (Sections~\ref{sec:main}-\ref{sec: forest}), we briefly survey the existing related work in Section~\ref{sec:related_work}, and provide the necessary background in Section~\ref{sec: background}. Finally, in Section~\ref{sec:exp}, we empirically investigate the efficiency and scalability of WFOMC with graph-theoretical constraints as a combinatorial framework, and investigate the increase in expressivity of MLNs with the presented constraints.

%%%%%%%%%%%%%%%%%%%%%%%%%%%%%%%%%%% RELATED WORK %%%%%%%%%%%%%%%%%%%%%%%%%%%%%%%%%%

\section{Related Work}
\label{sec:related_work}
WFOMC was initially formalized in \cite{PTP} and \cite{LIFTED_PI_KB_COMPLETION} for its applications to SRL frameworks like MLNs \cite{richardson2006markov} and PLPs \cite{PLP}. Initial works in WFOMC generalized \emph{knowledge compilation} techniques to FOL theories, providing  an algorithm for Symmetric-WFOMC over universally quantified FOL theories. The formal definition of lifted inference  as FOL theories admitting polynomial time weighted model counting w.r.t the domain cardinality was provided in \cite{domain_lifted_defintion}. This created a deep connection between the investigation of efficient probabilistic inference algorithms and the theoretical analysis of WFOMC. A significant advance in WFOMC techniques came in the form of a WFOMC-preserving skolemization procedure \cite{broeck2013}, which allowed for the complete two-variable fragment 
to admit polynomial time WFOMC. These works were followed by extensive complexity analysis of WFOMC \cite{Symmetric_Weighted}, providing a closed-form formula for WFOMC in the universally quantified fragment of FO$^2$. However, these results also showed that in the three-variable fragment of FOL there is a formula whose WFOMC cannot be computed in polynomial time, demonstrating that WFOMC is $\#$P$_1$-complete in general. This led to significant interest in expanding domain liftable fragments in other directions, for instance by adding functionality constraints \cite{kuusisto2018weighted} (i.e., a relation in the language is a function). A major expansion in domain-liftable fragments came in the form of domain-liftability of C$^2$ \cite{kuzelka2020weighted, AAAI_Sagar}. Recent results have focused on expanding the expressivity of C$^2$ with additional constraints like the linear order axiom \cite{Linear_Order_Axiom} and tree axiom \cite{Tree}.  The approaches proposed in this paper --- which proves domain-liftability of FO$^2$ extended with acyclicity, connectivity, forests and trees constraints --- are most closely related to \cite{Symmetric_Weighted, AAAI_Sagar, AIxIA}. We also exploit the results from \cite{kuzelka2020weighted} (especially for dealing with cardinality constraints) to expand our results to C$^2$. Our work is also related to recent results that expand WFOMC of C$^2$ with Tree axiom \cite{Tree}. However, the techniques introduced in this paper are different and more general than the ones used in  \cite{Tree}. This is because we are able to obtain trees as an instance of connected graphs with $n-1$ edges, which can be simply encoded in cardinality constraints. Finally, our work also draws significantly from combinatorics literature on DAGs \cite{robinson1973counting}, connected graphs \cite{Graphical_Enumeration} and forests \cite{Counting_Forests}. And it provides a new resource for tackling other problems of interest in combinatorics, like the enumeration of phylogenetic networks \cite{bienvenu_combinatorial_2022,cardona_counting_2020,fuchs_short_2021}.

%%%%%%%%%%%%%%%%%%%%%%%%%%%%%%%%%%% BACKGROUND %%%%%%%%%%%%%%%%%%%%%%%%%%%%%%%%

\section{Background} \label{sec: background}
This section overviews basic notation, assumptions and aspects of FOL and WFOMC, revisiting and reformulating existing works --- as relevant to our results --- on WFOMC in FO$^2$ and C$^2$. 
Additional necessary background is provided at the beginning of each section.

\subsection{Basic Notation} 
We use $[n]$ to denote the set of integers $\{1,\dots,n\}$. Whenever the set of integers $[n]$ is obvious from the context and $m\in [n]$ we will use $[\bar{m}]$ to represent the set $\{m+1, \dots, n\}$. Bold font letters (e.g.\  $\bk$) are used to denote integer vectors, and corresponding regular font letters (with an additional index) for the components of the vectors (e.g.\ $k_i$). Hence, we use $\bm{k} = \langle k_1,...,k_u \rangle$ to denote a vector of $u$ non-negative integers. Given an integer vector $\bk$, we use $|\bk|$ to denote $\sum_{i \in [u]}k_i$, i.e.\ the sum of all the components of $\bk$. 
We will also use the multinomial coefficients denoted by  $$\binom{|\bk|}{k_1,...,k_u} =  \binom{|\bk|}{\bk}\coloneqq  \frac{|\bk|!}{\prod_{i\in [u]} k_i !}$$
When summing two vectors $\bk' = \langle k'_1,...,k'_u \rangle $ and $\bk''= \langle k''_1,...,k''_u \rangle$, we always intend element-wise sum, i.e., $\bk' + \bk'' =  \langle k'_1+k''_1,...,k'_u+k''_u \rangle $.

\subsection{First-Order Logic} 
We assume a function-free FOL language $\mathcal{L}$ defined by a set of variables $\mathcal{V}$ and a set of
relational symbols $\mathcal{R}$. We write $R/k$ to denote that the relational symbol $R$ has arity $k$. For $(x_1,...,x_k) \in \mathcal{V}^{k}$ and $R/k \in \mathcal{R}$, we call $R(x_1,...,x_k)$ an \emph{atom}. A \emph{literal} is an atom or its negation. A \emph{formula} is formed by connecting atoms with boolean operators ($\neg, \lor$ and $\land $) and quantifiers of the form $\exists x_i.$ (existential quantification) and $\forall x_i.$ (universal quantification), using FOL syntax rules. The \emph{free} variables of a formula are those that are not bound by a quantifier. We write $\Phi(x_1,\dots x_k)$ to denote a formula whose free variables are $\{x_1,\dots,x_k\}$. An FOL formula with no free-variables is called a \emph{sentence}. Hence, a sentence is denoted by a capital Greek letter (e.g. $\Psi$). The sentences in $\mathcal{L}$ are interpreted over a set of constants called the \emph{domain}. The set of ground atoms (resp. ground literals) are obtained by replacing all the variables in the atoms (resp. literals) of $\mathcal{L}$ with domain constants. Hence, given a predicate $R/k$, and a domain $\Delta$ of size $n$, we have $n^{k}$ ground atoms of the form $R(a_1,\dots,a_k)$, where $(a_1\dots a_k) \in \Delta^{k}$. An \emph{interpretation} $\omega$, on a finite \emph{domain} $\Delta$,  is a truth assignment to  all the ground atoms. We assume Herbrand semantics \cite{Herbrand_Logic}, hence $\omega$ is an interpretation or a model of $\Psi$ if $\omega \models \Psi$ under Herbrand semantics. Given a literal (resp. ground literal) $l$, we use $pred(l)$ to denote the relational symbol in $l$. For a subset $\mathrm{\Delta'}$ of the domain $\Delta$, we use $\omega \downarrow \mathrm{\Delta'}$ to denote the partial interpretation induced on $\mathrm{\Delta'}$. Hence, $\omega \downarrow \mathrm{\Delta'}$ is an interpretation over the ground atoms containing only the domain elements in $\mathrm{\Delta'}$. Finally, we use $\omega_{R}$ to represent the partial interpretation of $\omega$ restricted to the relation $R$. When $R$ is binary, $\omega_{R}$ can be seen as a directed graph in which there is an edge from a domain element $c$ to a domain element $d$ whenever $\omega\models R(c,d)$. We can also restrict $\omega_R$ to undirected graphs by axiomatizing $R$ to be anti-reflexive and symmetric, by adding the axiom $\forall x. \neg R(x,x) \land \forall xy. R(x,y) \rightarrow R(y,x)$.

\begin{example}
    \label{ex: projection}
    Let us have a language with only the two relational symbols $R$ and $B$, both of arity $2$.  We represent an interpretation $\omega$ as a multi-relational directed graph, where a pair of domain elements $(c,d)$ has a red (resp.\ blue) directed edge from $c$ to $d$ if and only if $R(c,d)$ (resp.\ $B(c,d)$) is true in $\omega$. Let us take for example the following interpretation $\omega$ on $\Delta = [4]$:
\noindent
\begin{center}
\begin{tikzpicture}[node distance={13mm}, thick, main/.style = {draw, circle}] 
        \node[main] (1) {$2$}; 
        \node[main] (2) [left of=1] {$1$}; 
        \node[main] (3) [right of=1] {$3$}; 
        \node[main] (4) [right of=3] {$4$};  
        \draw[red, ->,line width = 2pt] (1) -- (2); 
        \draw[blue,->,line width = 2pt] (1) to [out=135,in=90,looseness=1.5] (4);
        \draw[red,->,line width = 2pt] (3) to [out=180,in=270,looseness=6] (3);
        \draw[blue,->,line width = 2pt] (3) -- (4); 
        \end{tikzpicture}
    \end{center}
\vspace{-1em}
\noindent Then, $\omega' = \omega \downarrow [2]$ and $\omega'' = \omega \downarrow [\bar{2}]$ are given respectively as
\noindent
\begin{center}
    \begin{tikzpicture}[node distance={13mm}, thick, main/.style = {draw, circle}] 
            \node[main] (1) {$2$}; 
            \node[main] (2) [left of=1] {$1$}; 
            \node[main] (3) [right = 4cm of 2] {$3$}; 
            \node[main] (4) [right of=3] {$4$};  
            \draw[red,->,line width = 2pt] (1) -- (2);
            \node[fill=none,align=center]{\hskip 8em and};
            \draw[red,->,line width = 2pt] (3) to [out=190,in=270,looseness=5] (3); 
             
            \draw[blue,->,line width = 2pt] (3) -- (4); 
            \end{tikzpicture}
\end{center}
\vspace{-1em}
 On the other hand, $\omega_{R}$ is obtained by projecting on the predicate $R$ and is given as
\begin{center}
    \begin{tikzpicture}[node distance={13mm}, thick, main/.style = {draw, circle}] 
            \node[main] (1) {$2$}; 
            \node[main] (2) [left of=1] {$1$}; 
            \node[main] (3) [right of=1] {$3$}; 
            \node[main] (4) [right of=3] {$4$};  
            \draw[red, ->,line width = 2pt] (1) -- (2); 
            \draw[red,->,line width = 2pt] (3) to [out=180,in=270,looseness=6] (3);
            \end{tikzpicture}
        \end{center}

\vspace{-1em}
\end{example}

We will also exploit the following proposition about purely universally quantified FOL sentences.

\begin{proposition} 
\label{prop: formula on subset}
 Let $\Psi$ be a first order logic sentence such that $\Psi$ is of the form $\forall \bx. \Phi(\bx)$, where $\bx = x_1, ..., x_k$ represents the free variables in $\Phi(\bx)$, and $\Phi(\bx)$ is quantifier-free. If $\omega$ is an interpretation over a domain $\Delta$, and $\omega \models \Psi$, then $\omega \downarrow \Delta' \models \Psi$ for all $\Delta' \subseteq \Delta$. 
\end{proposition}
\begin{proof} Under Herbrand semantics, $\omega \models \forall \bx. \Phi(\bx)$ can be equivalently written as $\omega \models \bigwedge_{\mathbf{c} \in \Delta^{k}}\Phi(\mathbf{c})$. Hence, $\omega \models \bigwedge_{\mathbf{c} \in \Delta'^{k}}\Phi(\mathbf{c})$, for any $\Delta' \subseteq \Delta$. Now, $\omega \downarrow \Delta'$ is the truth assignment to all the atoms on the domain $\Delta'$. Hence, $\omega \downarrow \Delta' \models \bigwedge_{\mathbf{c} \in \Delta'^{k}}\Phi(\mathbf{c})$. Therefore, $\omega \downarrow \Delta' \models \forall \bx. \Phi(\bx)$. 
\end{proof}

Proposition \ref{prop: formula on subset} allows us to split an interpretation over disjoint subsets of the domain while preserving satisfaction of a universally quantified FOL sentence. We will also deal with expanding interpretations from two disjoint subsets of the domain. Given disjoint sets of domain constants $\Delta'$ and $\Delta''$, we use $\Delta = \Delta' \uplus \Delta''$ to denote the fact that $\Delta$ is the union of the two disjoint sets.  If $\omega'$ is an interpretation on $\Delta'$ and $\omega''$ is an interpretation on $\Delta''$, then we use $\omega' \uplus \omega''$ to denote the partial interpretation on $ \Delta' \uplus \Delta''$ obtained by interpreting the ground atoms over $\Delta'$ as they are interpreted in $\omega'$ and the ground atoms over $\Delta''$ as they are interpreted in $\omega''$. The ground atoms involving domain constants from both $\Delta'$ and $\Delta''$ are left un-interpreted in $\omega' \uplus \omega''$. We illustrate this point in the following:

\begin{example}
    \label{ex: extension}
    Let us have an $\mathrm{FOL}$ language with only one relational symbol $R$, of arity 2. Let $\Delta = [3]$, $\Delta' = [2]$ and $\Delta'' = [\bar{2}] = \{3\}$. 
    We represent an interpretation $\omega$ as a directed graph where each pair $(c, d)$ of domain elements has a red edge from $c$ to $d$ if and only if $R(c,d)$ is true in $\omega$.
    Let us have the following two interpretations $\omega'$ and $\omega''$ on the domains $[2]$ and $[\bar{2}]$, respectively:
    \begin{center}
        \begin{tikzpicture}[node distance={13mm}, thick, main/.style = {draw, circle}] 
                \node[main] (1) {$2$}; 
                \node[main] (2) [left of=1] {$1$}; 
                \node[main] (3) [right = 4cm of 2] {$3$}; 
                \draw[red,->,line width = 2pt] (1) -- (2);
                \node[fill=none,align=center]{\hskip 8em and};
                \draw[red,->,line width = 2pt] (3) to [out=190,in=270,looseness=5] (3); 
                 
                \end{tikzpicture}
        \vspace{-2em}
        \begin{align*}
            \quad\quad \omega'  \hspace{8em}  \omega''
        \end{align*}
    \end{center}
We now create a partial interpretation $\omega' \uplus \omega''$ as follows:
\begin{center}
\begin{tikzpicture}[node distance={13mm}, thick, main/.style = {draw, circle}] 
    \node[main] (1) {$2$}; 
    \node[main] (2) [left of=1] {$1$}; 
    \node[main] (3) [right = 2cm of 2] {$3$}; 
    \draw[red,->,line width = 2pt] (1) -- (2);
    \draw[red,->,line width = 2pt] (3) to [out=190,in=270,looseness=5] (3); 
    \draw[gray, dotted, line width = 2pt] (1) to [out=90,in=90,looseness=0.7] (3);
    \draw[gray, dotted, line width = 2pt] (2) to [out=90,in=90,looseness=0.7] (3);
     
    \draw [dashed] ($0.42*(1)+0.42*(3) + (0,1)$) -- ($0.42*(1)+0.42*(3) + (0,-1)$);
    \end{tikzpicture}
\end{center}
The gray dotted lines \begin{tikzpicture}
    \draw[gray, dotted, line width = 2pt] (0,0) -- (0.5,0); 
\end{tikzpicture} represent the fact that $R(1,3)$, $R(3,1)$, $R(2,3)$ and $R(3,2)$ are not interpreted in $\omega' \uplus \omega''$. A possible extension of $\omega' \uplus \omega''$ is given as
\begin{center}
    \begin{tikzpicture}[node distance={13mm}, thick, main/.style = {draw, circle}] 
        \node[main] (1) {$2$}; 
        \node[main] (2) [left of=1] {$1$}; 
        \node[main] (3) [right = 2cm of 2] {$3$}; 
         
        \draw[red,->,line width = 2pt] (1) -- (2);
        \draw[red,->,line width = 2pt] (3) to [out=190,in=270,looseness=5] (3); 
        \draw[red, ->, line width = 2pt] (1) to [out=90,in=110,looseness=0.8] (3);
         
        \draw [dashed] ($0.42*(1)+0.42*(3) + (0,1)$) -- ($0.42*(1)+0.42*(3) + (0,-1)$);
        \end{tikzpicture}
    \end{center}
where $R(2,3)$ is interpreted to be true and $R(1,3)$, $R(3,1)$ and $R(3,2)$ are interpreted to be false. We can see that  $\omega' \uplus \omega''$ can be extended in $2^{4}$ ways.
\end{example}

\subsubsection{\texorpdfstring{FO$^2$}{FO²} and its Extensions} 

FO$^2$ is the fragment of FOL with only two variables. An extension of FO$^2$ is obtained by introducing \emph{counting quantifiers} of the form $\exists^{=k}$ (there exist exactly $k$), $\exists^{\geq k}$ (there exist at least $k$) and $\exists^{\leq k}$ (there exist at most $k$). This extended fragment is denoted by C$^2$ \cite{COUNTING_REF}. Cardinality constraints are constraints on the cardinality of predicates in an interpretation \cite{kuzelka2020weighted}. For example, $\omega \models \Phi \land (|R| \geq 5)$ iff  $\omega \models \Phi$ and the sum of the ground atoms in $\omega$, with the predicate $R$,  that are interpreted to be true is at least $5$.

\subsubsection{Types and Tables}

We will use the notion of $1$-types, $2$-type, and $2$-tables as presented in \cite{kuusisto2018weighted,ECML_PROJ}. A $1$-type is a maximally consistent conjunction of literals containing only one variable. For example, in a language with only the relational symbols $U/1$ and $R/2$, both ${U(x)\land R(x,x)}$ and ${U(x)\land \neg R(x,x)}$ are examples of 1-types with variable $x$. A 2-table is a maximally consistent conjunction of literals containing exactly two distinct variables. Extending the previous example, both $R(x,y)\land \neg R(y,x)\land (x \neq y)$ and $R(x,y)\land R(y,x) \land (x \neq y )$ are instances of 2-tables. We assume an arbitrary order on the 1-types and 2-tables. Hence, we use ${i(x)}$ (resp.\ $i(y)$) to denote the $i^{th}$ 1-type with variable $x$ (resp.\ with variable $y$), and $ {l(x,y)}$ to denote the $l^{th}$ 2-table. A $2$-type is a formula of the form ${i(x)\land j(y) \land l(x,y)}$, and we use ${ijl(x,y)}$ to represent it. In a given interpretation $\omega$, we say that a domain constant $c$ realizes the $i^{th}$ 1-type if $ {\omega \models i(c)}$, that a pair of domain constants $(c,d)$ realizes the $l^{th}$ 2-table if $ {\omega \models l(c,d)}$ and that $(c,d)$ realizes the 2-type $ {ijl(x,y)}$ if ${\omega \models ijl(c,d)}$. We will use $u$ to denote the number of 1-types and $b$ to denote the number of 2-tables in a given FOL language. For instance, in the language with the relational symbols $U/1$ and $R/2$, we have that  $u =2^{2}$ and $b = 2^{2}$.

\begin{definition}[1-type Cardinality Vector] An interpretation $\omega$ is said to have the 1-type cardinality vector $\bk = \langle k_1,\dots,k_u \rangle$ if for all $i \in [u]$, it has $k_i$ domain elements $c$ such that $\omega \models i(c)$, where $i(x)$ is the $i^{th}$ 1-type. If $\omega$ has 1-type cardinality vector $\bk$, then we say that $\omega \models \bk$. 
\end{definition}

Notice that in a given interpretation $\omega$ each domain element realizes exactly one 1-type. Hence, given a 1-type cardinality vector $\bk$, $|\bk|$ is equal to the domain cardinality. Furthermore, for a given $\bk$ and a fixed pair of 1-types $i$ and $j$, where $i \leq j$, there are $k_ik_j$ pairs of domain constants $(c,d)$ such that $\omega \models i(c) \land j(d)$.  Similarly, for a given $\bk$ and a 1-type $i$, there are $\binom{k_i}{2}$ unordered pairs of distinct domain constants $(c,d)$ such that $\omega \models i(c) \land i(d)$.

\subsection{Weighted First Order Model Counting} 
In WFOMC as defined in equation \eqref{eq: WFOMC}, we assume that the weight function $\w$ does not depend on individual domain constants, which implies that $\w$ assigns the same weight to two interpretations which are isomorphic under a permutation of domain elements. Hence, for a domain $\Delta$ of size $n$ we can equivalently use $[n]$ as our domain.  Furthermore, as common in literature, we will focus on a special class of weight functions, namely \emph{symmetric weight functions}, defined as follows:

\begin{definition}[Symmetric Weight Function]
    \label{def: symm}
    
    Let $\mathcal{G}$ be the set of all ground atoms and $\mathcal{R}$ the set of all relational symbols in a given FOL language. 
    A symmetric weight function associates two real-valued weights, $w: \mathcal{R} \rightarrow \mathbb{R}$ and $\bar{w}: \mathcal{R} \rightarrow \mathbb{R}$,  to each relational symbol in the language. The weight of an interpretation $\omega$ is then defined as follows:
    \begin{equation}
        \label{eq: symmweight}
        \w(\omega) \coloneqq \prod_{\substack{ \omega \models g \\ g \in \mathcal{G} \\}}w(pred(g)) \prod_{\substack{ \omega \models \neg g\\ g \in \mathcal{G} }}\bar{w}(pred(g)).
    \end{equation} 
    We use $(w,\bar{w})$ to denote a symmetric weight function.
\end{definition}

Many techniques \cite{broeck2013, kuzelka2020weighted, AAAI_Sagar} for WFOMC of a sentence $\Phi$  require WFOMC-preserving reductions. That is, constructing another sentence $\Phi'$ and a new weight function $(w',\bar{w}')$ such that
$$\wfomc(\Phi, (w,\bar{w}),n) = \wfomc(\Phi',(w',\bar{w}'),n).$$
A key property desired from a WFOMC-preserving reduction is that adding new formulas to it should not invalidate the reduction. Reductions satisfying this requirement are called \emph{modular}. We formalize this in the following definition:

\begin{definition} \cite{broeck2013} A reduction $(\Phi, w,\bar{w})$ to $(\Phi', w',\bar{w}')$, where $\Phi$ and $\Phi'$ are two sentences and $(w,\bar w)$ and $(w',\bar{w}')$ are two weight functions, is modular if
    \begin{equation}
        \label{eq: modularity}
        \wfomc(\Phi \land \Lambda,(w,\bar{w}),n) = \wfomc(\Phi' \land \Lambda,(w',\bar{w}'),n)  
    \end{equation}  
     for any sentence $\Lambda$.
\end{definition}
    
Intuitively, modular reductions are sound under the presence of another sentence $\Lambda$, and any new sentence $\Lambda$ does not invalidate a modular reduction.

For the rest of the paper, whenever referring to weights, we intend symmetric weights. Hence, we denote $\wfomc(\Phi,(w,\bar{w}), n)$ without explicitly mentioning the weights, i.e. as $\wfomc(\Phi, n)$. 
We will often compute the WFOMC of interpretations with a fixed 1-type cardinality vector, and with a slight abuse of notation denote it as $\wfomc(\Phi,\bk)$. Formally,
\begin{equation}
    \label{eq: k_decomposed_new}
    \wfomc(\Phi,\bk) := \sum_{\omega \models \Phi \land \bk} \w(\omega)
\end{equation}

\subsubsection{WFOMC in \texorpdfstring{FO$^2$}{FO²}}
We revisit  WFOMC for universally quantified FO$^2$ formulas, i.e.\ formulas of the form $\forall xy. \Phi(x,y)$, where $\Phi(x,y)$ is quantifier-free. We define $\Phi(\{x,y\})$ as 
$$\Phi(x,x)\land \Phi(x,y)\land \Phi(y,x)\land \Phi(y,y)\land( x \neq y)$$
We also define the notion of consistent 2-types as follows:

\begin{definition}
    \label{def: 2-type_consistency}
    Given an FO$^2$ sentence $\forall xy. \Phi(x,y)$, where $\Phi(x,y)$ is quantifier-free, we say that a $2$-type $ijl(x,y)$ is consistent with $\forall xy. \Phi(x,y)$ if
    \begin{align}
      \label{2_type_consistency}
      ijl(x,y) \models \Phi(\{x,y\})
    \end{align}  

    \end{definition}
\noindent
  Note that the entailment in equation \eqref{2_type_consistency} is checked by assuming a propositional language consisting of only the (constant free) atoms in the FO$^2$ language.

\begin{example}
    \label{ex:truth-assignment} Let $\Phi(x,y)\coloneqq (A(x) \land R(x,y)) \rightarrow A(y)$. The following is an example of a consistent $2$-type for the sentence $\forall xy. \Phi(x,y)$:
  \begin{align*}
   \tau(x,y) \coloneqq
   \neg A(x) \land R(x,x) \land \neg A(y) \land R(y,y) \land \neg R(x,y) \land R(y,x) \land (x\neq y)
  \end{align*}

\end{example}

A key idea in analyzing domain-liftability of universally quantified FO$^2$ formula  is that a pair of domain constants $(c,d)$ in an interpretation $\omega \models \forall xy. \Phi(x,y)$ can realize a 2-type $ijl(c,d)$ only if the 2-type is consistent with the formula $\forall xy. \Phi(x,y)$, i.e.\ only if $ijl(x,y) \models \Phi(\{x,y\})$. We formalize this intuition in the following proposition.

\begin{proposition}
 \label{prop: ext}
 Let $\forall xy. \Phi(x,y)$ be an FO$^2$ sentence where $\Phi(x,y)$ is quantifier-free. Then, $\omega \models \forall xy. \Phi(x,y)$ iff for any pair of distinct domain constants $(c,d)$, such that $\omega \models ijl(c,d)$, we have that $ijl(x,y)$ is consistent with $ \forall xy. \Phi(x,y)$, i.e.\ $ijl(x,y) \models \Phi(\{x,y\})$.
\end{proposition}

\begin{proof} 
    If $\omega \models \forall xy. \Phi(x,y)$ and $\omega \models ijl(c,d)$, then $\omega \models \forall xy. \Phi(x,y) \land ijl(c,d)$. Now, $ijl(c,d)$ is a complete truth assignment to the ground atoms containing only the domain constants $c$ or $d$ or both. Hence, $ \omega \models \forall xy. \Phi(x,y) \land ijl(c,d)$ only if $ijl(c,d) \models \Phi(\{c,d\})$, i.e.\ only if ${ijl(x,y) \models \Phi(\{x,y\})}$. 
    On the other hand, let $\omega$ be such that all pair of domain constants realize only the 2-types $ijl(x,y)$ consistent with $\forall xy. \Phi(x,y)$. It means that $\omega \models \Phi(\{c,d\})$ for all the pairs of domain constants $(c,d)$. Hence, $\omega \models \forall xy. \Phi(x,y)$.   
\end{proof}

To facilitate the treatment of WFOMC, we will now introduce some weight parameters associated with an FO$^2$ language. Consider an FO$^2$ language $\mathcal{L}$ with symmetric weight function $(w,\bar{w})$, and let $\mathcal{I}$ denote the set of atoms in $\mathcal{L}$ that contain only variables. We define the following weight parameters associated with each 1-type $i(x)$ and 2-table $l(x,y)$:

\begin{equation}\label{eq: w_i}
    w_i \coloneqq \prod_{\substack{i(x) \models  g \\ g \in \mathcal{I}}}w(pred(g)) \prod_{\substack{ i(x) \models \neg g \\ g \in \mathcal{I}}} \bar{w}(pred(g))
\end{equation}
and
\begin{equation}\label{eq: v_l}
    v_l \coloneqq  \prod_{\substack{l(x,y) \models  g \\ g \in \mathcal{I}}}w(pred(g)) \prod_{\substack{ l(x,y) \models \neg g \\ g \in \mathcal{I}}} \bar{w}(pred(g))
\end{equation}

We now present an analytical formula for WFOMC of FO$^2$ sentences of the form $\forall xy. \Phi(x,y)$ where $\Phi(x,y)$ is quantifier-free. Our presentation is based on the treatment proposed in \cite{Symmetric_Weighted}.

\begin{theorem}[reformulated from \cite{Symmetric_Weighted}]
    \label{thm: beam}
    Given an FO$^2$ sentence $\forall xy. \Phi(x,y)$ where $\Phi(x,y)$ is quantifier-free, 
    let us define, for any pair of $1$-types $(i,j)$, the quantity $r_{ij} \coloneqq \sum_{l\in[b]}n_{ijl}v_{l}$, with $n_{ijl}$ being $1$ if  $ijl(x,y) \models \Phi(\{x,y\})$ and $0$ otherwise. Then, the weighted model count of $\forall xy. \Phi(x,y)$ with $1$-types cardinality vector $\bk$ is given as follows:
    \begin{align}
        \label{eq: WFOMC_Beame}
        \wfomc(\forall xy.\Phi(x,y),\bk) =  \binom{|\bk|}{\bm{k}} \prod_{i\in [u]}w_i^{k_i}
        \prod_{\substack{{i\leq j \in[u]}}}\!\!\! r_{i j}^{\bk(i,j)} 
    \end{align} 
where $\bk(i,j)$ is defined as
\begin{align*}
    \label{kij}
  \bk(i,j) \coloneqq
  \begin{cases} 
      \frac{k_{i}(k_{i} - 1)}{2} & \text{if $i=j$} \\
       k_{i}k_{j} & \text{otherwise} \\
     \end{cases}
\end{align*} 

\end{theorem}

\begin{proof}
   In a $1$-type cardinality vector $\bk$, and $k_i$ represents the number of domain constants that realize the $1$-type $i$. Since any domain constant realizes one and only one $1$-type in any given interpretation, there are exactly $\binom{|\bk|}{\bk}$ ways of assigning $1$-types to $|\bk|$ domain constants. Suppose that a domain constant $c$ realizes the $i^{th}$ $1$-type for an interpretation $\omega \models \bk$; then $i(c)$ contributes to the weight of the interpretation  with the weight $w_i$, multiplicatively. Therefore, the contribution due to $1$-type realizations is given by $\prod_{i\in [u]}w_i^{k_i}$ for any interpretation $\omega \models \bk$. Summing over all such interpretations gives us the factor of $\binom{|\bk|}{\bk}\prod_{i\in [u]}w_i^{k_i}$.
    
    Now consider an interpretation $\omega$ and a pair of distinct domain constants $(c, d)$ such that $\omega \models i(c)\land j(d)$. Using Proposition \ref{prop: ext}, we know that $(c,d)$ realizes the 2-table $l(c,d)$ if and only if $ijl(x,y) \models \Phi(\{x,y\})$. Therefore, in an arbitrary interpretation $\omega$, the multiplicative weight contribution due to the realization of the $l^{th}$ 2-table by a pair of constants $(c,d)$ such that $\omega \models i(c) \land j(d)$ is given by $n_{ijl}v_{l}$. Also, each ordered pair of constants can realize exactly one and only one $2$-table. Hence, the sum of the weights of the possible $2$-table realizations of a pair of domain constants $(c,d)$ such that $i(c)$ and $j(d)$ are true is given as $r_{ij}=\sum_{l}n_{ijl}v_{l}$. Furthermore, given the $1$-type assignments $i(c)$ and $j(d)$, the ordered pair $(c,d)$ can realize $2$-tables independently of all other domain constants.  Since there are $\bk(i,j)$ possible such pairs, the contribution given by the realization of $2$-tables over all the interpretations $\omega \models \bk$ is $\prod_{\substack{{i\leq j \in[u]}}}%\!\!\!
    r_{i j}^{\bk(i,j)}$. 
    \end{proof}

Equation \eqref{eq: WFOMC_Beame} can be computed in polynomial time w.r.t.\ the domain cardinality $n$, and there are only polynomially many $\bk$ w.r.t.\ $n$. Hence, $\wfomc(\forall xy.\Phi(x,y),n)$, which is equal to 
\begin{equation}
\label{eq: WFOMC_full_beame}
\sum_{|\bk|=n}\wfomc(\forall xy.\Phi(x,y),\bk)
\end{equation}
can be computed in polynomial time w.r.t.\ domain size $n$. Furthermore, \cite{broeck2013} shows that any FOL formula with existential quantification can be modularly reduced to a WFOMC preserving universally quantified FO$^2$ formula, with additional new predicates and negative weights. Hence, showing that FO$^2$ is domain-liftable.

\begin{remark}
    Note that the number of distinct $\bk$ in equation \eqref{eq: WFOMC_full_beame} can be super-exponentially many w.r.t the number of symbols in the language. However, as common in domain-liftability literature, in this paper we are only concerned  with computational complexity of WFOMC w.r.t. the domain cardinality $n$. 
\end{remark}

\subsubsection{WFOMC in \texorpdfstring{C$^2$}{C²}}
\cite{AAAI_Sagar} and \cite{kuzelka2020weighted}  show that WFOMC in C$^2$ can be reduced to WFOMC in FO$^2$ with cardinality constraints. The key result that leads to domain liftability of C$^2$, is the following:

\begin{restatable}{theorem}{thmCardinality}(\cite{kuzelka2020weighted}, slightly reformulated)\label{th: cardinality}
  Let $\Phi$ be a first-order logic sentence and let $\Gamma$ be an arbitrary cardinality constraint. Then, $\wfomc(\Phi \land \Gamma,\bk)$ can be computed in polynomial time with respect to the domain cardinality, relative to the $\wfomc(\Phi,\bk)$ oracle. 
\end{restatable}

In order to prove Theorem \ref{th: cardinality}, the proof in \cite{kuzelka2020weighted} relies on polynomial interpolation. We provide an easier presentation of the proof using Gauss-elimination based polynomial interpolation in the appendix.

\begin{remark} 
    \label{rem: FOL_inexpressible_Card}
    In the proof presented in the appendix (and in \cite{kuzelka2020weighted}), the first-order definability of $\Phi$ is never invoked. This property has also been exploited for imposing cardinality constraints with tree axiom in \cite{Tree}.    
\end{remark}

Theorem \ref{th: cardinality} extends domain-liftability of any sentence $\Phi$ to its domain-liftability with cardinality constraints. We now move to domain-liftability in C$^2$:

\begin{theorem}[\cite{kuzelka2020weighted}]
    \label{kuzelka_C2}
     The fragment of first-order logic with two variables and counting quantifiers is domain-liftable.
\end{theorem}

 The key idea behind Theorem \ref{kuzelka_C2} is that WFOMC of a C$^2$ sentence $\Phi$ can be converted to a problem of WFOMC of an FO$^2$ sentence $\Phi'$ with cardinality constraints $\Gamma$ on an extended vocabulary with additional weights for the new predicates (the new predicates are weighted 1 or -1). We refer the reader to \cite{kuzelka2020weighted} and \cite{AAAI_Sagar} for the detailed treatment of Theorem \ref{kuzelka_C2}. However, for our purposes it is important to note that this transformation is modular.  The modularity of  the WFOMC procedure, as presented in \cite{kuzelka2020weighted}, has also been exploited to demonstrate domain-liftability of C$^2$ extended with Tree axiom \cite{Tree} and Linear Order axiom \cite{Linear_Order_Axiom}.
%%%%%%%%%%%%%%%%%%%%%%%%%%%%%%%% MAIN RESULTS %%%%%%%%%%%%%%%%%%%%%%%%%%%%%%%
\section{Main Approach: Counting by Splitting}\label{sec:main}
As shown in Example \ref{ex: extension}, the possible ways of \emph{merging} two interpretations $\omega'$ and $\omega''$, on disjoint domains $\Delta'$ and $\Delta''$,   can be efficiently computed. Moreover, additional constraints can be imposed on the way the interpretations are merged. Such constraints, for instance, may express statements like \say{for the binary predicate $R$, there is no $R$-edge going from $\Delta''$ to $\Delta'$}. Formally, we express this constraint as $\forall x \in \Delta' \,\, \forall y \in \Delta''.  \neg R(y,x)$. 

\begin{example} Among the $2^4$ possible extensions of $\omega'\uplus \omega''$ in Example \ref{ex: extension}, the following $2^{2}$  satisfy the additional constraint ${\forall x \in [2] \, \forall y \in [\bar{2}]. \neg R(y,x)}$:
\begin{center}
    \begin{tikzpicture}[node distance={13mm}, thick, main/.style = {draw, circle}] 
        \node[main] (1) {$2$}; 
        \node[main] (2) [left of=1] {$1$}; 
        \node[main] (3) [right = 2cm of 2] {$3$}; 
        \draw[red,->,line width = 2pt] (1) -- (2);
        \draw[red,->,line width = 2pt] (3) to [out=190,in=270,looseness=5] (3); 
        \draw [dashed] ($0.42*(1)+0.42*(3) + (0,1)$) -- ($0.42*(1)+0.42*(3) + (0,-1)$);
    \end{tikzpicture}
    \quad\quad\quad\quad
    \begin{tikzpicture}[node distance={13mm}, thick, main/.style = {draw, circle}] 
        \node[main] (1) {$2$}; 
        \node[main] (2) [left of=1] {$1$}; 
        \node[main] (3) [right = 2cm of 2] {$3$}; 
        \draw[red,->,line width = 2pt] (1) -- (2);
        \draw[red,->,line width = 2pt] (3) to [out=190,in=270,looseness=5] (3); 
        \draw[red, ->, line width = 2pt] (2) to [out=90,in=90,looseness=0.7] (3);
        \draw [dashed] ($0.42*(1)+0.42*(3) + (0,1)$) -- ($0.42*(1)+0.42*(3) + (0,-1)$);
    \end{tikzpicture}
    \\
    \begin{tikzpicture}[node distance={13mm}, thick, main/.style = {draw, circle}] 
        \node[main] (1) {$2$}; 
        \node[main] (2) [left of=1] {$1$}; 
        \node[main] (3) [right = 2cm of 2] {$3$}; 
        \draw[red,->,line width = 2pt] (1) -- (2);
        \draw[red,->,line width = 2pt] (3) to [out=190,in=270,looseness=5] (3); 
        \draw[red, ->, line width = 2pt] (1) to [out=90,in=110,looseness=0.7] (3);
        \draw [dashed] ($0.42*(1)+0.42*(3) + (0,1)$) -- ($0.42*(1)+0.42*(3) + (0,-1)$);
    \end{tikzpicture}
    \quad\quad\quad\quad
    \begin{tikzpicture}[node distance={13mm}, thick, main/.style = {draw, circle}] 
        \node[main] (1) {$2$}; 
        \node[main] (2) [left of=1] {$1$}; 
        \node[main] (3) [right = 2cm of 2] {$3$}; 
        \draw[red,->,line width = 2pt] (1) -- (2);
        \draw[red,->,line width = 2pt] (3) to [out=190,in=270,looseness=5] (3); 
        \draw[red, ->, line width = 2pt] (1) to [out=90,in=110,looseness=0.7] (3);
        \draw[red, ->, line width = 2pt] (2) to [out=90,in=90,looseness=0.7] (3);
        \draw [dashed] ($0.42*(1)+0.42*(3) + (0,1)$) -- ($0.42*(1)+0.42*(3) + (0,-1)$);
    \end{tikzpicture}
\end{center}
\end{example}

A key insight of the paper is that such merging can also be done for disjoint interpretations of a universally quantified FO$^2$ formula, while preserving the satisfaction of such formula. This is formalized in the following lemma.

\begin{lemma} \label{lem: count_split} Let $\Phi(x,y)$ and $\Theta(x,y)$  be quantifier-free $\mathrm{FO^2}$ formulas. Let $\omega'$ and $\omega''$ be interpretations satisfying $\forall xy. \Phi(x,y)$ on two disjoint domains $\Delta'$ and $\Delta''$ (respectively). We define $r_{ij} := \sum_{l\in[b]} n_{ijl}v_l$, where $n_{ijl}$ is $1$ if $ijl(x, y)\models \Phi(\{x, y\}) \land \Theta(x,y)$ and $0$ otherwise, and $v_l$ is defined in equation~\eqref{eq: v_l}. Then, the sum of the weights of the extensions $\omega$ of $\omega' \uplus \omega''$ such that
\begin{itemize}
    \item $\omega \models \forall xy. \Phi(x,y)$
    \item $\omega \models \forall x \in \Delta'\,\, \forall y \in \Delta''. \Theta(x,y)$
\end{itemize}is given as
\begin{equation}
\label{eq:lemma_count_split}
    \w(\omega')\w(\omega'') \prod_{i,j\in [u]} r^{k'_i k''_j}_{ij}
\end{equation}
where $k'_i$ and $k''_j$ are the number of domain constants realizing the $i^{th}$ and $j^{th}$ $1$-types in $\omega'$ and $\omega''$ respectively.
\end{lemma}
\begin{proof}
In order to obtain an interpretation $\omega$ from $\omega' \uplus \omega''$, we only need to extend $\omega' \uplus \omega''$ with interpretations of the ground-atoms containing pairs $(c,d) \in \Delta' \times \Delta''$. 
For a given pair $(c,d) \in \Delta' \times \Delta''$, let $i$ and $j$ be the $1$-types such that $\omega' \models i(c)$ and $\omega'' \models j(d)$.
Since we want $\omega$ to be a model of $\forall xy. \Phi(x,y)$, if $\omega\models l(c,d)$, then by Proposition \ref{prop: ext} we must have that  $ijl(c,d) \models \Phi(\{c,d\})$. 
Moreover, since we want $\omega$ to satisfy the condition $\forall x \in \Delta'\, \forall y \in \Delta''. \Theta(x,y)$, we must have that $ijl(c,d) \models \Theta(c,d)$.
Therefore, $(c,d)$ can realize the $l^{th}$ 2-table if and only if $n_{ijl}=1$. The multiplicative weight contribution to the weight of a given extension $\omega$ of $\omega' \uplus \omega''$ due to $(c,d)$ realizing the $l^{th}$ 2-table is given as $n_{ijl}v_l$. Furthermore, $(c,d)$ realizes the 2-tables mutually-exclusively. Also, the weight contribution of the 2-table realizations of $(c,d)$ only depends on the 1-types of $c$ and $d$ and is independent of all other domain constants. 
Hence, $r_{ij} = \sum_{l \in [b]}n_{ijl}v_l$ is the multiplicative weight contribution given by the interpretation of all the possible ground atoms containing a pair $(c,d) \in \Delta'\times \Delta''$ with $c$ realizing the $i^{th}$ $1$-type in $\omega'$ and $j$ realizing the $j^{th}$ $1$-type in $\omega''$. 
Since there are $k'_i$ domain elements $c$ realizing the $i^{th}$ $1$-type in $\omega'$, and $k''_j$ domain elements $d$ realizing the $j^{th}$ $1$-type in $\omega''$, the weight contribution of the interpretation of all the possible ground atoms containing a pair $(c,d) \in \Delta' \times \Delta''$  is given by $\prod_{i,j\in[u]} r_{ij}^{k'_i k''_j}$.
Furthermore, since $\omega'$ and $\omega''$ interpret two disjoint sets of ground atoms, their weights $\w(\omega')$ and $\w(\omega'')$ contribute independently to each $\w(\omega)$.  
\end{proof}

Lemma \ref{lem: count_split} provides us a method to compute the (weighted sum of the) number of ways to merge two FO$^2$ interpretations on two disjoint domains. We will now use Lemma \ref{lem: count_split} to compute weighted model count of FO$^2$ formulas, where the models are restricted to obey additional constraints (potentially inexpressible in FOL)  on disjoint subsets of the domain.

\begin{lemma}[Counting by Splitting]\label{lem: count_split_wmc}
Let $\Phi(x,y)$ and $\Theta(x,y)$ be quantifier-free FO$^2$ formulas. Let $\Psi' \coloneqq \forall xy. \Phi(x,y) \land axiom'$ and $\Psi'' \coloneqq \forall xy. \Phi(x,y) \land axiom''$, where $axiom'$ and $axiom''$ denote arbitrary constraints -- potentially inexpressible in first-order logic. Let $[n]$ be the domain, and let $m\leq n$ be a natural number. Let $\Psi_{[m]}$ denote the conjunction of the following four conditions on interpretations $\omega$:
\begin{enumerate}[C1:~~]
    \item $\omega\downarrow [m] \models axiom'$
    \item $\omega\downarrow [\bar{m}] \models axiom''$
    \item $\omega \models \forall x \in [m]\,\, \forall y \in [\bar m]. \Theta(x,y)$
     \item $\omega \models \forall xy. \Phi(x,y)$
\end{enumerate}
Then, the weighted model count of the interpretations $\omega$ satisfying both $\Psi_{[m]}$ and the cardinality vector $\bk$ is given as:
\begin{equation}\label{eq:lemma_Psi_m}
    \wfomc(\Psi_{[m]}, \bk) = \!\sum_{\substack{\bk' + \bk''=\bk \\ |\bk'|=m}} \!\!\! \wfomc(\Psi', \bk') \wfomc(\Psi'', \bk'')\!\! \prod_{i,j\in [u]} \!\! r^{k'_i k''_j}_{ij} 
\end{equation}
where  
$r_{ij} := \sum_{l\in[b]} n_{ijl}v_l$, and $n_{ijl}$ is $1$ if $ijl(x, y)\models \Phi(\{x, y\}) \land \Theta(x,y)$ and $0$ otherwise, and $v_l$ is as defined in equation~\eqref{eq: v_l}.
\end{lemma}

\begin{proof} 
Let $\omega\models \Psi_{[m]}$. Using condition C4 and Proposition \ref{prop: formula on subset}, we have that  $\omega\downarrow [m] \models \forall xy. \Phi(x,y)$. Furthermore, using condition C1, we have that $\omega\downarrow [m] \models axiom'$. Hence, $\omega\downarrow [m] \models \Psi'$, and similarly $\omega\downarrow [\bar{m}] \models \Psi''$. The WFOMC of $\Psi'$ on $[m]$ with 1-type cardinality vector $\bk'$ is $\wfomc(\Psi',\bk')$. Similarly, the WFOMC of $\Psi''$ on $[\bar{m}]$ with 1-type cardinality vector $\bk''$ is $\wfomc(\Psi'',\bk'')$. For a given pair of models $\omega'$ and $\omega''$ counted respectively in $\wfomc(\Psi',\bk')$ and $\wfomc(\Psi,\bk'')$, the weighted sum of extensions $\omega \models \Psi_{[m]}$ is given by equation~\eqref{eq:lemma_count_split}, due to Lemma~\ref{lem: count_split}. Summing over extensions of all such possible pairs of  $\omega'$ and $\omega''$, we get:
    \begin{equation}
        \label{eq: meta_wfomc_k'_k''}
        \wfomc(\Psi',\bk')\wfomc(\Psi'',\bk'' )\!\! \prod_{i,j\in[u]} \!\!r_{ij}^{k'_ik''_j}
    \end{equation}
    Expression \eqref{eq: meta_wfomc_k'_k''} gives us the WFOMC of the models $\omega$ such that  $\omega \models \Psi_{[m]}$, ${\omega \downarrow [m]  \models \bk'}$ and $\omega \downarrow [\bar{m}]  \models \bk''$. 
    To compute the WFOMC of the models $\omega$ such that $\omega \models \Psi_{[m]} \land \bk$, 
    we have to sum expression \eqref{eq: meta_wfomc_k'_k''} over all possible  
    decompositions of the $1$-type cardinality vector $\bk$ into $\bk'$ and $\bk''$, where $|\bk'| = m$.
\end{proof}

The techniques developed in this paper will use counting by splitting with different instantiations of  $axiom'$, $axiom''$ and $\Theta(x,y)$.
The following table summarizes how we will instantiate them when providing the algorithms for WFOMC with DAG, connected graph and forest constraints (as given respectively in Definitions \ref{def: Acyclic graph}, \ref{def: Connceted Graph} and \ref{def:forest_axiom}, and making use of Remarks \ref{rem:DAG}, \ref{rem: symmetric_con} and \ref{rem:symmetric_forest}).

\begin{table}[H]
\label{tab:summary}
    \centering
    \begin{tabular}{c|c c c}
        Constraint & $axiom'$ & $axiom''$ & $\Theta(x,y)$ \\
        \hline
        $DAG(R)$ & $\forall xy.\neg R(x,y)$ & $DAG(R)$ & $\neg R(y,x)$  \\
        $Connected(R)$ &  $Connected(R)$ & $\top$ & $\neg R(x,y)$ \\
        $Forest(R)$ & $Tree(R)$ & $Forest(R)$ & $\neg R(x,y)$
    \end{tabular}
    \caption{Axioms used for counting by splitting for different constraints.}
\end{table}
%%%%%%%%%%%%%%%%%%%%%%%%%%%%%%%%%%%% DAG AXIOMS %%%%%%%%%%%%%%%%%%%%%%%%%%%%%%%
\section{WFOMC with DAG Axiom}
\label{sec: DAG}
A \emph{Directed Acyclic Graph (DAG)} is a directed graph such that  starting from an arbitrary node $i$ and traversing an arbitrary path along directed edges we would never arrive back at node $i$. We present a recursive formula based on the principle of inclusion-exclusion for WFOMC of an FO$^2$ sentence, say $\Phi$, where a special predicate, say $R$, is axiomatized to be a DAG, i.e.\ we count the models $\omega \models \Phi$ such that $\omega_{R}$ represents a DAG. We begin by revisiting the principle of inclusion-exclusion and the recursive formula for counting DAGs on $n$ nodes, as presented in \cite{robinson1973counting}. We identify key observations that lead to the formula for counting DAGs and exploit analogous ideas for WFOMC with the DAG axiom. We then expand the DAG axiom with \emph{source} and \emph{sink} constraints. Our results also shed a new light on counting phylogenetic networks \cite{bienvenu_combinatorial_2022,cardona_counting_2020,fuchs_short_2021} -- a widely investigated problem in combinatorics and mathematical biology.

\subsection{Principle of Inclusion-Exclusion}
Given a set of finite sets $\{A_i\}_{i\in [n]}$, let $A_{J}:= \bigcap_{j\in J}A_j$ for any subset $J\subseteq [n]$. The \emph{principle of inclusion-exclusion (PIE)} states that:
\begin{equation}
    \label{P_IE}
    \Big|\bigcup_{i}A_i\Big| = \sum_{\emptyset \neq J \subseteq [n]} (-1)^{|J|+1} \big| A_{J} \big|
\end{equation} 
If the cardinality of the intersections $A_J$ only depends on the cardinality of $J$, then the formula can be simplified. In this case, all $A_J$'s with $|J| = m$ have cardinality equal to $|A_{[m]}|$. Since for any $m\in [n]$ there are $\binom{n}{m}$ sets $A_J$ with $|J|=m$, equation~\eqref{P_IE} reduces to:
\begin{equation}
    \label{P_IE_Symm}
    \Big|\bigcup_{i}A_i\Big| = \sum_{m=1}^{n} (-1)^{m+1} \binom{n}{m} \big |A_{[m]} \big |
\end{equation}

    The principle of inclusion-exclusion can be easily extended to the case when the sets $A_i$ contain weighted FOL interpretations and we want the weighted sum of all the interpretations in $\bigcup_{i}A_i$. Indeed, if $\w(A_i)$ denotes the weighted sum of all the interpretations in $A_i$, then the PIE reduces to: 
    \begin{equation}
        \label{P_IE_w}
        \w\Big( \bigcup_{i}A_i \Big) = \sum_{\emptyset \neq J \subseteq [n]} (-1)^{|J|+1} \w(A_{J})
    \end{equation}
Analogously, when $\w(A_{J})=\w(A_{J'})$ for all $J, J'$ with the same cardinality, then:
\begin{equation}
    \label{P_IE_Symm_w}
    \w\Big( \bigcup_{i}A_i \Big) = \sum_{m=1}^{n} (-1)^{m+1} \binom{n}{m} \w(A_{[m]})
\end{equation}

\subsection{Counting Directed Acyclic Graphs}
\label{sec: DAG_Count}
We now derive the formula for counting DAGs as presented in \cite{robinson1973counting}.
 Without loss of generality, we assume the set of nodes to be $[n]$. For each $i\in [n]$, let $A_i$ be the set of DAGs on $[n]$ for which the node $i$ has indegree zero. Since every DAG has at least one node with indegree zero, we have that the total number of DAGs $a_n$ is equal to $|\bigcup_{i\in [n]}A_i|$. The set of DAGs such that all nodes in $J \subseteq [n]$ have indegree zero is given by $A_{J}:= \bigcap_{j \in J} A_j$. The cardinality of $A_J$ depends only on the cardinality of $J$, and not on the individual elements in $J$. Hence, we can assume w.l.o.g that $J = [m]$, where $m = |J|$. We now derive a method for computing $\big | A_{[m]} \big |$. We make the following key observations:

\begin{itemize}
    \item Observation 1. If $\omega \in A_{[m]}$, then there are no edges between the nodes in $[m]$, as otherwise a node in $[m]$ will have a non-zero indegree. In other words, only directed edges from $[m]$ to $[\bar{m}]$ or within $[\bar m]$ are allowed.
    \item Observation 2. If $\omega \in A_{[m]}$, then the subgraph of $\omega$ restricted to $[\bar{m}]$, i.e.\ $\omega \downarrow [\bar{m}]$, is a DAG. And the subgraph of $\omega$ restricted to $[m]$ is just an empty graph, i.e.\ the set of isolated nodes $[m]$ with no edges between them. 

    \item Observation 3. Any DAG on $[\bar{m}]$ can be extended to $2^{m(n-m)}$ DAGs in $A_{[m]}$. This is because DAGs in $A_{[m]}$ have no edges between the nodes in $[m]$; they only have outgoing edges from $[m]$ to $[\bar{m}]$. For extending a given DAG on $[\bar{m}]$ to a DAG in $A_{[m]}$, we have two choices for each pair of nodes in $[m]\times [\bar{m}]$: we can either draw an out-going edge from $[m]$ to $[\bar{m}]$ or not. Hence, there are $2^{|[m]\times [\bar{m}]|} = 2^{m(n-m)}$ ways to extend a given DAG on $[\bar{m}]$ to a DAG in $A_{[m]}$.  
\end{itemize}

The number of possible DAGs on $[\bar{m}]$ is $a_{n-m}$. Due to Observation 3, we have that $A_{[m]}$ has $2^{m(n-m)} a_{n-m}$ DAGs obtained by extending the DAGs on $[\bar m]$. Furthermore, due to Observation 1 and Observation 2, these are all the possible DAGs in $A_{[m]}$. Hence, $|A_{[m]}| =  2^{m(n-m)} a_{n-m}$. Now we can repeat this argument for any $m$-sized subset of $[n]$: if $|J|=|J'| = m$, then $|A_{J}| = |A_{J'}| =  2^{m(n-m)} a_{n-m}$. Using the principle of inclusion-exclusion as given in equation \eqref{P_IE_Symm}, we have that:

\begin{equation}
\label{eq: Count_DAG}    
a_n = \sum_{m=1}^{n}(-1)^{m+1}\binom{n}{m}2^{m(n-m)}a_{n-m}
\end{equation}
Notice that, after replacing $n-m$ with $l$, the equation becomes:
\begin{equation}
    \label{eq: Count_DAG_l}    
    a_n = \sum_{l=0 }^{n-1}(-1)^{n-l+1}\binom{n}{l}2^{l(n-l)}a_{l}
\end{equation}
The change of variable allows us to write a bottom-up algorithm for counting DAGs, given in Algorithm \ref{alg:algorithm_DAG}. Based on this algorithm, it can be seen that $a_n$  can be computed in polynomial time w.r.t $n$. Indeed, the $\mathbf{for}$ loop in line 4 runs $n$ times, and in line 5 we compute $a_i$ (and store it as $A[i]$) using equation \eqref{eq: Count_DAG_l}, which requires only polynomially many  operations in $n$.

\begin{algorithm}[tb]
    \caption{Number of DAG on $n$ nodes}
    \label{alg:algorithm_DAG}
    \begin{algorithmic}[1]
    \State \textbf{Input}: $n$
    \State \textbf{Output}: $a_n$
        \State $A[0] \gets 1$ 
        \For{$i=1$ to $n$}
        \State $A[i] \gets \sum_{l=0}^{i-1}(-1)^{i-l+1}\binom{i}{l}2^{l(i-l)}A[l]$ 
        \EndFor\\
        \textbf{return} $A[n]$
        \end{algorithmic}
\end{algorithm}

\subsection{WFOMC with DAG Axiom}
\begin{definition}
    \label{def: Acyclic graph}
    Let $R$ be a binary predicate. We say that an interpretation $\omega$ is a model of $Acyclic(R)$, and write $\omega \models Acyclic(R)$, if $\omega_R$\, forms a DAG.
\end{definition}

\begin{remark}
\label{rem:DAG}
    Our goal is to compute the WFOMC of\,\,$\Psi \coloneqq \forall xy.\Phi(x,y) \land Acyclic(R)$, where $\Phi(x,y)$ is a quantifier-free FO$^2$ formula. By definition, this entails that $\omega_R$ \,forms an antireflexive relation of $R$. Hence, we can assume without loss of generality that $\forall xy.\Phi(x,y) \models \forall x. \neg R(x,x)$.
\end{remark}
\noindent

W.l.o.g., we assume that the domain is $[n]$. We will now redefine $\Psi_{[m]}$ for the purposes of WFOMC with DAG axiom.

\begin{definition}
    \label{def: atleast_m_shorter}
    Let $\Psi \coloneqq \forall xy. \Phi(x,y) \land Acyclic(R)$, where $\Phi(x,y)$ is a quantifier-free FO$^2$ formula  
    such that $\forall xy. \Phi(x,y) \models \forall x. \neg R(x,x)$. For any $m \leq n$, we say that an interpretation $\omega$ is a model of $\Psi_{[m]}$, and write $\omega \models \Psi_{[m]}$, if  $\omega$ is a model of\, $\Psi$ on $[n]$ and the domain elements $[m]$ have indegree zero in $\omega_R$.
\end{definition}
\begin{comment}
In the following, we show that Definition \ref{def: atleast_m_shorter} is indeed an instantiation of $\Psi_{[m]}$ as introduced in Lemma \ref{lem: count_split_wmc}.
\begin{lemma}%[$\Psi_{[m]}$]
    \label{lem: atleast_m_DAG}
    An interpretation $\omega$ is a model of $\Psi_{[m]}$ (according to the previous definition) if and only if $\omega$ satisfies conditions C1-C4 in  Lemma \ref{lem: count_split_wmc}, where we instantiate 
    $axiom'$, $axiom''$ and $\Theta(x,y)$ as follows:
    \begin{itemize}
        \item $axiom' \coloneqq \forall xy. \neg R(x,y)$
        \item $axiom'' \coloneqq Acyclic(R)$
        \item $\Theta(x,y) \coloneqq \neg R(y,x)$
    \end{itemize} 
\end{lemma}
\begin{proof} 
In Definition \ref{def: atleast_m_shorter}, since nodes in $[m]$ have zero in-degree, we have that $\omega \downarrow [m] \models axiom'$. Since, subgraph of an acyclic graph is acyclic, we have that $\omega \downarrow [\bar{m}] \models axiom''$. Since $[m]$ has indegree zero, hence, there can be no edge from $[\bar{m}]$ to $[m]$, i.e., $\omega \models \forall x \in [m]\,\, \forall y \in [\bar m]. \Theta(x,y)$. C4 is satisfied by construction. Finally, any   \red{Should be an if and only if.}
\end{proof}
\end{comment}
It can be shown that $\Psi_{[m]}$ in Definition \ref{def: atleast_m_shorter} corresponds to an instantiation of  $\Psi_{[m]}$ in Lemma~\ref{lem: count_split_wmc}, where we instantiate  $axiom'$, $axiom''$ and $\Theta(x,y)$ as follows:
    \begin{itemize}
        \item $axiom' \coloneqq \forall xy. \neg R(x,y)$
        \item $axiom'' \coloneqq Acyclic(R)$
        \item $\Theta(x,y) \coloneqq \neg R(y,x)$
    \end{itemize}
A formal proof of this correspondence is in the appendix (Lemma \ref{lem: atleast_m_DAG}). 
With this specific instantiation of $\Psi_{[m]}$, we can use Lemma \ref{lem: count_split_wmc} and get the following proposition. 
\begin{proposition}
    \label{prop: atleast_weighted}
        Let $\Psi \coloneqq \forall xy. \Phi(x,y) \land Acyclic(R)$ and $\Psi' \coloneqq \forall xy. \Phi(x,y) \land \neg R(x,y)$, where $\Phi(x,y)$ is a quantifier-free FO$^2$ formula, such that $\forall xy. \Phi(x,y) \models \forall x. \neg R(x,x)$. Then: 
    \begin{equation}
        \label{eq: k_k'_m}
        \wfomc(\Psi_{[m]},\bk) =
          \!\sum_{\substack{\bk' + \bk''=\bk \\ |\bk'|=m}} \!\!\! \wfomc(\Psi', \bk') \wfomc(\Psi, \bk'')\!\! \prod_{i,j\in [u]} \!\! r^{k'_i k''_j}_{ij} 
    \end{equation}
        where $r_{ij} := \sum_{l}n_{ijl}v_{l}$, and $n_{ijl}$ is $1$ if  $ijl(x,y) \models \Phi(\{x,y\}) \land \neg R(y,x)$ and $0$ otherwise. 
    \end{proposition}

\noindent
Similarly to equation \ref{eq: Count_DAG}, we can use $\wfomc(\Psi_{[m]},\bk)$ and the principle of inclusion-exclusion for computing the WFOMC for $\Psi \coloneqq \forall xy. \Phi(x,y) \land Acyclic(R)$.

\begin{proposition}Let $\Psi \coloneqq \forall xy. \Phi(x,y) \land Acyclic(R)$, where $\Phi(x,y)$ is a quantifier-free FO$^2$ formula, such that $\forall xy. \Phi(x,y) \models \forall x. \neg R(x,x)$. Then:
    \begin{equation}
        \label{eq: DAG_Acyclic}
        \wfomc(\Psi ,\bk) = \sum_{m=1}^{|\bk|}(-1)^{m + 1}\binom{|\bk|}{m}\wfomc(\Psi_{[m]},\bk)
    \end{equation} 
    
\end{proposition}

\begin{proof} The proof idea is very similar to the case for counting DAGs as given in equation \eqref{eq: Count_DAG}. W.l.o.g., we are assuming the domain to be $[n]$, so $|\bk| = n$. Let $A_{i}$ be the set of models $\omega$ of  $\Psi$, such that $\omega$ has 1-type cardinality $\bk$ and the domain element $i$ has zero $R$-indegree in $\omega_R$. Since every DAG has at least one node with zero $R$-indegree, our goal is to compute  $\w(\cup_{i\in [n]}A_i)$. For any $J\subseteq [n]$, let $A_{J}\coloneqq  \bigcap_{j\in J}A_j$. For any $J$ and $J'$ such that $|J| = |J'|$, we always have that $|A_{J}| = |A_{J'}|$. Hence, using the principle of inclusion-exclusion as given in equation \eqref{P_IE_Symm_w}, we have that:
    \begin{equation}
        \label{e: PIE_Acyclic}
        \wfomc(\Psi ,\bk)  = \sum_{m=1}^{|\bk|} (-1)^{m+1} \binom{n}{m}\w(A_{[m]}) 
    \end{equation}   
Now, $A_{[m]}$ is the set of models such that domain elements in $[m]$ have zero $R$-indegree. Hence, $\w(A_{[m]}) = \wfomc(\Psi_{[m]},\bk)$. 
\end{proof}
We make a change of variable in equation \eqref{eq: DAG_Acyclic} --- similar to the one from \eqref{eq: Count_DAG} to \eqref{eq: Count_DAG_l} --- by replacing $m$ with $|\bk|-l$. We obtain the following equation:

\begin{align}
    \label{eq: DAG_Acyclic_FO_l}
        \wfomc(\Psi ,\bk)=
    \sum_{l=0}^{|\bk|-1}(-1)^{ |\bk| -l + 1}\binom{|\bk|}{l}\wfomc(\Psi_{[|\bk|-l]},\bk)
\end{align} 
We provide  pseudocode for evaluating equation \eqref{eq: DAG_Acyclic_FO_l} in Algorithm \ref{alg:algorithm_FO_DAG}, namely WFOMC-DAG. We now  analyze how WFOMC-DAG works, and show that it runs in polynomial time with respect to domain cardinality $|\bk| = n$.  

WFOMC-DAG takes as input $\Psi = \forall xy. \Phi(x,y) \land Acyclic(R)$ and $\bk$ and it returns $\wfomc(\Psi,\bk)$. In line $3$, an array $A$ with $u$ indices is initiated, and $A[\mathbf{0}]$ is assigned the value  $1$, where $\mathbf{0}$ corresponds to the $u$-dimensional zero vector. The for loop in lines $5-7$ incrementally computes $\wfomc(\Psi,\bp)$, where the loop runs over all $u$-dimensional integer vectors $\bp$, such that $p_i \leq k_i$, in lexicographical order. The number of possible $\bp$ vectors is bounded by $n^{u}$. Hence, the  for loop in line 5 runs at most $n^{u}$ iterations. In line 6, we compute $\wfomc(\Psi, \bp)$ as given in equation \eqref{eq: DAG_Acyclic_FO_l}. Also in line 6,  the function $\overline{\wfomc}(\Psi_{[m]},\bp)$ --- that computes $\wfomc(\Psi_{[m]},\bp)$ --- is called at most $|\bp|-1$ times, which is bounded above by $n$. $A[\bp]$ stores the value $\wfomc(\Psi,\bp)$.  In the function $\overline{\wfomc}(\Psi_{[m]},\bs)$, the number of iterations in the for loop (line 12) is bounded above by $n^{2u}$. And $\wfomc(\Psi',\bs')$ is an FO$^2$ WFOMC problem, again computable in polynomial time. Hence, the algorithm WFOMC-DAG runs in polynomial time w.r.t domain cardinality. Notice that since loop 5-7 runs in lexicographical order, the $A[\bs'']$ required in the function $\overline{\wfomc}(\Psi_{[m]},\bs)$ are always already stored in $A$.

\begin{algorithm}[tb]
    \caption{WFOMC-DAG}
    \label{alg:algorithm_FO_DAG}
    \begin{algorithmic}[1]
    \State \textbf{Input}: $\Psi, \bk$
    \State \textbf{Output}: $\wfomc(\Psi,\bk)$
        \State $A[\mathbf{0}] \gets 1$ \Comment{$A$ has $u$ indices}
        \State \Comment{$\mathbf{0} = \langle 0,...,0 \rangle$ }
        \For{$\mathbf{0} < \bp \leq \bk$ where $ \bp \in \mathbb{N}_{0}^{u}$} \Comment{Lexical order} 
        \State ${\!A[\bp] \gets \!\sum_{l=0}^{|\bp|-1}(-1)^{|\bp|-l+1}\cdot \binom{|\bp|}{l}\cdot{\overline{\wfomc}}(\!\Psi_{[|\bp|-l]},\bp)}$ 
        \EndFor\\
        \textbf{return} $A[\bk]$ 
        \State
        \Function{$\overline{\wfomc}$}{$\Psi_{[m]}$, $\bs$} \Comment{Equation \eqref{eq: k_k'_m}}
        \State $S = 0$
        \For{$\bs' + \bs'' = \bs$ and $|\bs'| = m$}
        \State $S \gets S + \wfomc(\Psi',\bs')\cdot A[\bs''] \cdot \prod_{i,j\in[u]} r_{ij}^{s'_is''_j}$
        \EndFor
        \State \Return $S$
        \EndFunction
        \end{algorithmic}
\end{algorithm}

There are only polynomially many $\bk$ w.r.t domain cardinality. Hence, computing $\wfomc(\Psi,\bk)$ over all possible $\bk$ values, we can compute $\wfomc(\Psi, n)$ in polynomial time w.r.t domain cardinality.
Using the modular WFOMC-preserving Skolemization process as provided in \cite{broeck2013}, we can extend this result to prove the domain liftability of the entire FO$^2$ fragment, with DAG axiom.

Using Theorem \ref{th: cardinality} and Remark \ref{rem: FOL_inexpressible_Card}, we can also extend domain-liftability of FO$^2$, with DAG axiom and cardinality constraints.
Finally, since WFOMC of any C$^2$ formula can be modularly reduced to WFOMC of an FO$^2$ formula with cardinality constraints \cite{kuzelka2020weighted}, we have the following theorem:

\begin{theorem}
    \label{thm: counting_acyclic}
    Let $\Psi \coloneqq \Phi \land Acyclic(R)$, where $\Phi$ is a C$^2$ sentence. $\wfomc(\Psi,n)$ can be computed in polynomial time with respect to the domain cardinality.
\end{theorem}

Our results expand and cover previously investigated problems in combinatorics literature, such as  counting the number of DAGs with a fixed number of edges and nodes \cite{DAG_COUNT_ARCS}. Such results in the combinatorics literature provide a simple benchmark and sanity-check for our methodology. Hence, we will often check if model-counts for constraints axiomatized using our methodology align with the counts reported in the On-Line Encyclopedia of Integer Sequences (OEIS) \cite{oeis} --- an open-source collection of sequences, enumerating solutions to various counting problems. We illustrate this in the following example. 
\begin{example}\label{ex:DAGs_edges}
    Suppose we want to count the number of DAGs with $n$ labeled nodes and $d$ edges, as can be done through \cite{DAG_COUNT_ARCS}. Clearly, this can be encoded in FOL with DAG constraint as follows 
    \begin{equation}\label{eq: DAGs_edges}
        \fomc(Acyclic(R) \land |R|=d, n)
    \end{equation} 
    for an FOL language with only the binary relation $R$.
    Indeed, with our implementation of Algorithm \ref{alg:algorithm_DAG}, the FOMC in \eqref{eq: DAGs_edges}, for different values of $d$ and $n$, leads to the same sequence as A081064 of the OEIS \cite{oeis}.
    
This (and later examples) show that our implementation provides a simple method to enumerate structures respecting interesting combinatorial constraints directly from their logical encoding --- without the need to explicitly derive a different formula for each constraint.
\end{example}

\subsection{Source and Sink}

In the following, we show that extending C$^2$ with a DAG axiom allows us to easily express sources and sinks. We then provide some examples from combinatorics where this is useful.

\begin{definition}
    \label{def: source_sink}
    Let $\Phi$ be a first order sentence, possibly containing a binary relation $R$, a unary relation $Source$ and a unary relation $Sink$. We say that an interpretation $\omega$ is a model of  $\,\Phi \land Acyclic(R, Source, Sink)$ if:
    \begin{itemize}
        \item $\omega$ is a model of $\Phi \land Acyclic(R)$,
        \item the sources of the DAG represented by $\omega_{R}$ are interpreted to be true in $\omega_{Source}$,
        \item the sinks of the DAG represented by $\omega_{R}$ are interpreted to be true in $\omega_{Sink}$. 
    \end{itemize}
\end{definition}
The $Source$ and the $Sink$ predicates can allow us to encode constraints like $\exists^{=k}x.Source(x)$ or $\exists^{=k}x.Sink(x)$. See the example below.

\begin{theorem}Let $\Psi := \Phi \land Acyclic(R,Source,Sink)$, where $\Phi$ is a C$^2$ sentence. $\wfomc(\Psi,n)$ can be computed in polynomial time with respect to the domain cardinality.
\end{theorem}
\begin{proof}
  The sentence  $\Psi$ can be equivalently written as: 
  \begin{align}
    \begin{split}
    &\Phi \land Acyclic(R) \\
    & \land \forall x. Source(x) \leftrightarrow \neg \exists y. R(y,x) \\
    & \land \forall x. Sink(x) \leftrightarrow \neg \exists y. R(x,y)\\
\end{split}
  \end{align}
which is a C$^2$ sentence with a DAG constraint, for which Theorem \ref{thm: counting_acyclic} applies.
\end{proof}

\begin{example}\label{ex:DAGs_source_sink}
    For an FOL language made only of the binary relation $R$,
     \begin{equation}\label{eq: DAGs_edges_one_source}
        \fomc(Acyclic(R, Source, Sink) \land |R|=d \land |Source|=1, n)
    \end{equation} 
    counts the number of DAGs with $n$ labeled nodes, $d$ edges and a unique source. 
    Similarly, 
    \begin{equation}\label{eq: DAGs_one_source}
        \fomc(Acyclic(R, Source, Sink) \land |Source|=s, n)
    \end{equation} 
    counts the number of DAGs with $n$ labeled nodes and $s$ sources, while
    \begin{equation}\label{eq: DAGs_one_source_one_sink}
        \fomc(Acyclic(R, Source, Sink) \land |Source|=s \land |Sink| = s' , n)
    \end{equation} 
    counts the number of DAGs with $n$ labeled nodes, $s$ sources and $s'$ sinks.
    Indeed, with our axiomatization and implementation, we were able to compute the following sequences from OEIS:
    \begin{itemize}
        \item A350487 (DAGs with $n$ labeled nodes, $d$ edges and $1$ source);
        \item A361718 (DAGs with $n$ labeled nodes and $s$ sources), A003025 (DAGs with $n$ labeled nodes and $1$ source), A003026 (DAGs with $n$ labeled nodes and $2$ sources);
        \item A165950 (DAGs with $n$ labeled nodes, $1$ source and $1$ sink).
    \end{itemize}
    
    \noindent These counting problems were also investigated in \cite{counting_DAG_Source_Sink, dags_sources_sinks_mathoverlow}.
\end{example}

\begin{example}\label{ex:phylogenetic}
Binary phylogenetic networks \cite{mansouri_counting_2022, fuchs_counting_2021, pons2021combinatorial, mcdiarmid_counting_2015, cardona_counting_2020} are defined as DAGs consisting only of the following types of nodes:
\begin{itemize}
    \item one and only one source, which must have outdegree $2$;
    \item leaves, which are sinks with indegree $1$;
    \item tree nodes, which are nodes with indegree $1$ and outdegree $2$;
    \item reticulation nodes, which are nodes with indegree $2$ and outdegree $1$.
\end{itemize}
\noindent Along with their many subclasses, they are commonly used for representing the evolutionary relationships among a group of taxa, taking into account reticulate events like hybridization, horizontal gene transfer and recombination \cite{huson_rupp_scornavacca_2010}.
Counting binary phylogenetic networks is an interesting open problem in combinatorics \cite{mansouri_counting_2022}. Indeed, many recent studies are devoted to providing asymptotic estimates \cite{mcdiarmid_counting_2015, mansouri_counting_2022}, or the exact counting when the number of reticulation nodes is limited \cite{mansouri_counting_2022}, or estimates and exact counting for related networks, like tree-child and normal networks \cite{fuchs2022counting, mcdiarmid_counting_2015, pons2021combinatorial, fuchs_short_2021, chang_et_al:LIPIcs.AofA.2022.5, bienvenu_combinatorial_2022, fuchs_counting_2021, fuchs2021asymptotic, cardona_counting_2020, bouvel2020counting}. 

Notice that the definition of binary phylogenetic networks is expressible in C$^2$ with the acyclicity axiom:  
\begin{equation}
\begin{split}
   \forall x. [ (  S & ource(x) \land \exists^{=2}y. R(x,y)) \\
    \lor & (Sink(x) \land \exists^{=1}y. R(y,x)) \\
    \lor & (\exists^{=1}y. R(y,x) \land \exists^{=2}y. R(x,y))\\
    \lor & (\exists^{=2}y. R(y,x) \land \exists^{=1}y. R(x,y)) ]\\
    \land  | So & urce| = 1 \land Acyclic(R, Source, Sink)
\end{split}
\end{equation}
Hence, FOMC with acyclicity axiom can provide a way to count binary phylogenetic networks with a fixed number of nodes in polynomial time. Moreover, this approach is flexible enough to allow to count also other related classes like tree-child networks, and to count binary phylogenetic networks with any constraint expressible in $\mathrm{C}^2$. However, enumerating constraints with counting quantifiers requires an FOMC-preserving reduction to $\mathrm{FO}^2$ with cardinality constraints, with many additional predicates (see \cite{kuzelka2020weighted, AAAI_Sagar}), and the computational complexity scales super-exponentially with respect to the number of predicates. The number of additional predicates required in this case poses significant scalability issues.
\end{example}

%%%%%%%%%%%%%%%%%%%%%%%%%%%%%%%%%%%%% CONNECTIVITY %%%%%%%%%%%%%%%%%%%%%%%%%%%%%

\section{WFOMC with Connectivity Axiom}
\label{sec: connected}

A \emph{connected graph} is an undirected graph such that for any pair of nodes $i$ and $j$ there exists a path connecting the two nodes. We present a recursive formula for WFOMC of an FO$^2$ sentence, say $\Phi$, where a special predicate, say $R$, is axiomatized to be a connected graph, i.e.\ we count the models $\omega \models \Phi$ such that $\omega_{R}$ represents a connected graph. We begin by revisiting  the recursive formula for counting connected graphs on $n$ nodes. Much in the spirit of the previous section, we make observations that allow efficient counting of connected graphs and exploit them for WFOMC with connectivity axiom.

\subsection{Counting Connected Graphs} 
\label{sec: Con_Count}

In a given undirected graph, a \emph{connected component} is a subgraph that is not part of any larger connected subgraph. In a \emph{rooted graph}, one node is labeled in a special way, and called the \emph{root} of the graph. Given a rooted graph, we call its connected component containing the root as the \emph{rooted-connected component}. We now present a recursive way to count the number $c_n$ of connected graphs on $n$ nodes, as done in \cite{Graphical_Enumeration}. The base case is $c_1 = 1$.

\begin{proposition}
    \label{lem: k_conn_root_general}
    The number of rooted graphs on $[n]$ with an $m$-sized rooted-connected component is given as:
    \begin{equation}
        \label{eq: K_conn_general}
         \binom{n}{m}\cdot m\cdot c_m \cdot 2^{\binom{n-m}{2}}
    \end{equation}
    where $c_m$ is the number of connected graphs on $m$ nodes.
\end{proposition}
\begin{proof}
    Let $\omega$ be a rooted graph such that $\omega \downarrow [m]$ forms a rooted-connected component. Since $\omega \downarrow [m]$ is a connected component, there can be no edges between $[m]$ and $[\bar{m}]$. The number of possible connected graphs on $[m]$ is given by $c_{m}$. Also, in $\omega$ any node in $[m]$ can be chosen to be the root. Hence, the number of ways in which $\omega \downarrow [m]$ can be a rooted-connected component is $m \cdot c_m$. Since $\omega \downarrow [m]$ is a connected-component, there can be no edges between $[m]$ and $[\bar{m}]$, and $\omega \downarrow [\bar{m}]$ can be any $n-m$ sized graph. Hence, $2^{\binom{n-m}{2}}$ subgraphs can be realized on $[\bar{m}]$. Since subgraphs on $[m]$ and $[\bar{m}]$ are realized independently, the total number of graphs on $[n]$ such that $[m]$ is a rooted-connected component is given by $m\cdot c_m \cdot 2^{\binom{n-m}{2}}$.
    These arguments can be repeated for any rooted graph on $[n]$ with a rooted-connected component of size $m$. Since there are $\binom{n}{m}$ ways of choosing such subsets, we get formula \eqref{eq: K_conn_general}.
\end{proof}

Summing up equation \eqref{eq: K_conn_general} over all $m$, for $1 \leq m \leq n$, we get the total number of rooted graphs. This is the base of the proof of the following proposition, which gives us a way to recursively count the number of connected graphs on $n$ nodes. 

\begin{proposition}[\cite{Graphical_Enumeration}]
    \label{prop: conn_graphs}
    For any $m$, let $c_m$ be the number of connected graphs on $m$ nodes. Then, the following holds:
    \begin{equation}
    \label{eq: connected_graph}
    c_n = 2^{\binom{n}{2}} - \frac{1}{n}\sum_{m=1}^{n-1} \binom{n}{m}\cdot m\cdot c_m \cdot 2^{\binom{n-m}{2}} 
\end{equation}
\end{proposition}
\begin{proof}
Any rooted graph on $[n]$ has a rooted-connected component of some size $m$, where $1 \leq m \leq n$. Hence, 
\begin{equation}
    \sum_{m=1}^{n} \binom{n}{m}\cdot m\cdot c_m \cdot 2^{\binom{n-m}{2}} 
\end{equation}
is counting all the rooted graphs on $[n]$, whose number is $n \cdot 2^{\binom{n}{2}}$.
This gives us the following equation, where we have simply rewritten the summation in a different way:
\begin{equation}
    \label{eq: inter_mediate_2}
    n \cdot 2^{\binom{n}{2}} =  n \cdot c_n + \sum_{m=1}^{n-1} \binom{n}{m}\cdot m\cdot c_m \cdot 2^{\binom{n-m}{2}} 
\end{equation}
Clearly, this equation can be equivalently rewritten as equation \eqref{eq: connected_graph}.
\end{proof}

\subsection{WFOMC with Connectivity Axiom}

\begin{definition}
    \label{def: Connceted Graph}
    Let $R$ be a binary predicate. An interpretation $\omega$ is a model of
    $Connected(R)$ if 
    \begin{itemize}
        \item $\omega_R$ \/ forms a symmetric and antireflexive relation of $R$, and 
        \item $\omega_R$ \/ forms a connected graph.  
    \end{itemize}

\end{definition}

\begin{remark}
\label{rem: symmetric_con}
    Our goal is to compute WFOMC of $\Psi \coloneqq \forall xy.\Phi(x,y) \land Connected(R)$, where $\Phi(x,y)$ is a quantifier-free FO$^2$ formula interpreted on the domain $[n]$. By definition, this entails that $\omega_R$ forms a symmetric and antireflexive relation of $R$. Hence, we can assume without loss of generality that: 
    \begin{equation}\label{eq:wlog_symmmetric}
    \forall xy.\Phi(x,y) \models \forall x. \neg R(x,x) \land \forall xy. R(x,y) \rightarrow R(y,x)
    \end{equation}
\end{remark}

\begin{definition}
    \label{def: atleast_m_connected}
    Let $\Psi \coloneqq \forall xy.\Phi(x,y) \land Connected(R)$, where $\Phi(x,y)$ is a quantifier-free FO$^2$ formula interpreted on the domain $[n]$, such that \eqref{eq:wlog_symmmetric} holds. For any $m \leq n$, we say that an interpretation $\omega$ is a model of\, $\Psi_{[m]}$, and write $\omega \models \Psi_{[m]}$, if  $\omega$ is a model of $\Psi$ on $[n]$ and $\omega_R \downarrow [m]$ is a connected component in $\omega_R$.
   
\end{definition}

It can be shown that Definition \ref{def: atleast_m_connected} corresponds to an instantiation of $\Psi_{[m]}$ as introduced in Lemma \ref{lem: count_split_wmc} (see appendix Lemma \ref{lem: atleast_m_connected} for a formal proof), where 
\begin{itemize}
        \item $axiom' \coloneqq Connected(R)$
        \item $axiom'' \coloneqq \top$
        \item $\Theta(x,y) \coloneqq \neg R(x,y)$
\end{itemize} 
\noindent
Using this instantiation of $\Psi_{[m]}$ and Lemma~\ref{lem: count_split_wmc}, we have the following proposition. 

\begin{proposition}
\label{prop_con_m}
Let $\Psi \coloneqq \forall xy. \Phi(x,y) \land \C$ and $\Psi''\coloneqq \forall xy. \Phi(x,y)$, where $\Phi(x,y)$ is a quantifier-free FO$^2$ formula such that \eqref{eq:wlog_symmmetric} holds. Then:

\begin{equation}
    \label{eq: con_k_k'_m}
    \wfomc(\Psi_{[m]},\bk) =  
    \!\sum_{\substack{\bk' + \bk''=\bk \\ |\bk'|=m}} \!\!\! \wfomc(\Psi, \bk') \wfomc(\Psi'', \bk'')\!\! \prod_{i,j\in [u]} \!\! r^{k'_i k''_j}_{ij} 
\end{equation}
    where 
    $r_{ij} := \sum_{l}n_{ijl}v_l$, and $n_{ijl}$ is $1$ if  ${ijl(x,y) \models \Phi(\{x,y\}) \land \neg R(x,y)}$ and $0$ otherwise. 
\end{proposition}
Similar to Proposition \ref{lem: k_conn_root_general}, we will now compute the WFOMC for interpretations with rooted connected-components of size $m$.
\begin{proposition}
    \label{prop_con_C_general}
    Let $\Psi \coloneqq \forall xy. \Phi(x,y) \land Connected(R)$ be
    a sentence interpreted over $[n]$, where $\Phi(x,y)$ is a quantifier-free FO$^2$ formula such that \eqref{eq:wlog_symmmetric} holds. 
    Then the WFOMC of all the models $\omega \models \forall xy. \Phi(x,y)$ with a rooted-connected component of size $m$ is given as: 
    \begin{equation}
        \label{eq: PsiC_general}
        \binom{n}{m} \cdot m \cdot \wfomc(\Psi_{[m]},\bk)
    \end{equation}
    \end{proposition}
\begin{proof} The proof idea is similar to the one of Proposition \ref{lem: k_conn_root_general}.  There are $\binom{n}{m}$ ways of choosing an $m$-sized subset $C$ in $[n]$. Given such a set $C$, the number of models $\omega \models \forall xy. \Phi(x,y)$ with $C$ as a connected component w.r.t $R$ is $\wfomc(\Psi_{[m]},\bk)$. Finally, if we also allow a node in $C$ to be distinguished as a root, then we have $m$ ways of choosing the root.  
\end{proof}

Summing up equation \eqref{eq: PsiC_general} over all $m$, for $1 \leq m \leq n$, we get the total number of interpretations $\omega \models \forall xy. \Phi(x,y)$, where $\omega_{R}$ is a rooted graph. This is the base of the proof of the following proposition, which gives us a way to recursively compute the WFOMC of $\Psi \coloneqq \forall xy. \Phi(x,y) \land \C$.

\begin{proposition}\label{prop: WFOMC_CON} Let $\Psi \coloneqq \forall xy. \Phi(x,y) \land \C$ and $\Psi'' \coloneqq \forall xy. \Phi(x,y)$,
where $\Phi(x,y)$ is a quantifier-free FO$^2$ formula such that \eqref{eq:wlog_symmmetric} holds. Then: 
\begin{align}
    \label{eq: WFOMC_CON_final}
        \wfomc(\Psi ,\bk) = 
        \wfomc(\Psi'',\bk) - \frac{1}{n}\sum_{m=1}^{n-1}\binom{n}{m}\cdot m \cdot \wfomc(\Psi_{[m]},\bk) 
\end{align}
\end{proposition}

\begin{proof} The proof idea is similar to the one of Proposition \ref{prop: conn_graphs}. 
We notice that
\begin{align}
\label{eq: WFOMC_CON_new}
    \sum_{m=1}^{n}\binom{n}{m}\cdot m \cdot \wfomc(\Psi_{[m]},\bk)
\end{align}
sums the WFOMC of all the models $\omega$ of $\forall xy. \Phi(x,y)$ for which $\omega_{R}$ is a simple graph with an $R$-rooted-connected component of size $m$, where $1\leq m \leq n$. But any rooted graph has a rooted-connected component of some size $m$, where $1\leq m \leq n$. Hence, equation \eqref{eq: WFOMC_CON_new} computes the weighted sum of all models of $\forall xy. \Phi(x,y)$ where $\omega_R$ is a rooted graph. This is  equal to $n$ times the WFOMC of $\forall xy. \Phi(x,y)$, because we have $n$ choices for assigning a root in each model. Thus, we get the following equation
\begin{align}
    \label{eq: WFOMC_CON_split}
    \begin{split}
        &n\cdot \wfomc(\forall xy. \Phi(x,y) ,\bk) = \\
        &n\cdot \wfomc(\Psi_{[n]},\bk) + \sum_{m=1}^{n-1}\binom{n}{m}\cdot m \cdot \wfomc(\Psi_{[m]},\bk) 
    \end{split}
\end{align}
and since $\Psi_{[n]}$ is equivalent to $\Psi$, it can be rewritten as equation \eqref{eq: WFOMC_CON_final}.
\end{proof}

Using Proposition \ref{prop_con_m} and Proposition \ref{prop: WFOMC_CON} we can derive a recursive algorithm, very similar to Algorithm~\ref{alg:algorithm_FO_DAG}, for computing  $\wfomc(\Psi, \bk)$. This can be achieved by replacing line 6 and line 13 of Algorithm~\ref{alg:algorithm_FO_DAG} with the formulas coming from equations \eqref{eq: con_k_k'_m} and \eqref{eq: WFOMC_CON_final}. That is, replacing  line 6 and line 13, with the following operations:
\begin{algorithmic}[1]
\setcounter{ALG@line}{5}
    \State ${A[\bp] \gets \wfomc(\Psi'', \bp) - \frac{1}{|\bp|}\sum_{m=1}^{|\bp|-1} \binom{|\bp|}{m}\cdot m \cdot {\overline{\wfomc}}(\!\Psi_{[m]},\bp)}$ 
\end{algorithmic}
and
\begin{algorithmic}[1]
\setcounter{ALG@line}{12}
    \State $S \gets S + A[\bs'] \cdot \wfomc(\Psi'',\bs'') \cdot \prod_{i,j\in[u]} r_{ij}^{s'_is''_j}$   
\end{algorithmic}

The analysis of the modified algorithm for WFOMC with a connectivity constraint (see Algorithm \ref{alg:algorithm_CON}) is provided in the appendix. Almost all of the tractability analysis for Algorithm~\ref{alg:algorithm_FO_DAG} transfers identically to Algorithm \ref{alg:algorithm_CON}, showing that the algorithm is tractable. Hence giving us the following theorem:

\begin{theorem}
    \label{thm: con_c2}
    Let $\Psi \coloneqq \Phi \land \C$, where $\Phi$ is a C$^2$ formula. Then $\wfomc(\Psi,n)$ can be computed in polynomial time with respect to the domain cardinality.
\end{theorem}
The following examples show possible applications of our results in the realm of combinatorics.

\begin{example}\label{ex:connected_edges}
    For a FOL language made only of the binary relation $R$,
     \begin{equation}\label{eq: connected_tree}
        \fomc(Connected(R) \land |R|= 2d, n)
    \end{equation} 
    counts the number of connected graphs with $n$ labeled nodes and $d$ edges. For it to be non-zero, $d$ must be between $n-1$ (trees) and $\binom{n}{2}$ (complete graph). Such counting problem is studied in chapter 6.3 of \cite{bona2015handbook}.
    With our implementation, we were able to get the array A062734 of OEIS, which is also equivalent to A123527 and A343088. Its diagonals also give the following sequences: A000272 (number of trees on $n$ nodes), A057500 (number of connected unicyclic graphs on $n$ nodes), A061540, A061541, A061542, A061543, A096117, A061544 A096150, A096224, A182294, A182295, A182371.
\end{example}

\begin{example}\label{ex:3-colored}
    Suppose we want to count the number of connected graphs with $n$ labeled nodes, each one colored by one of three colors, with the condition that adjacent nodes must have different colors. This problem was already studied and solved in \cite{colored_connected}, where such number is denoted by $m_n(3)$. 
    We consider a FOL language with a relational symbol $R/2$ for representing the connected graph, and three relational symbols $A/1, B/1, C/1$ representing the colors. The fact that one and only one color is assigned to each node can be encoded with the sentence
    \begin{equation}
    \begin{split}
        \Psi_1 \coloneqq \forall x. ( A & (x) \lor B(x) \lor C(x)) \\
        \land & \neg (A(x) \land B(x)) \land \neg (A(x) \land C(x)) \land \neg (B(x) \land C(x)) 
    \end{split}
    \end{equation} 
    while the fact that adjacent nodes must have different colors can be encoded with the sentence
    \begin{equation}
    \begin{split}
        \Psi_2 \coloneqq \forall xy. &( A(x) \land R(x,y) \to \neg A(y)) \\ 
        \land &(B(x) \land R(x,y) \to \neg B(y)) \\
        \land &(C(x) \land R(x,y) \to \neg C(y))
    \end{split}
    \end{equation}
    Hence, the number of colored graphs that we want can be computed as
    \begin{equation}\label{eq: 3_colored}
        \fomc(\Psi_1 \land \Psi_2 \land Connected(R), n)
    \end{equation}
    and with our implementation we were able to compute sequence A002028 of OEIS. 
\end{example}

%%%%%%%%%%%%%%%%%%%%%%%%%%%%%%%%%%%%%%%%%%%FOREST_TREE%%%%%%%%%%%%%%%%%%%%%%%%%%%%%%%%%%

\section{Trees and Forests}\label{sec: forest}

In this section, we investigate the domain liftability of C$^2$ where one of the relational symbols is axiomatized to be a \emph{tree} or a \emph{forest}. Tree axioms have been previously investigated in  \cite{Tree}. However, we provide a completely different method of WFOMC for both directed and undirected tree axioms. Furthermore, we also demonstrate domain liftability of both directed and undirected forest axioms --- an open problem raised in the conclusion of \cite{Tree}.

\subsection{WFOMC with Directed Tree and Directed Forest Axioms}
 
\emph{Directed trees} are DAGs with exactly one node with indegree zero and with all the other nodes having indegree 1. Similarly, \emph{directed forests} are DAGs such that every node has indegree at most 1. (Equivalently, directed forests are DAGs such that each connected component is a directed tree.)

\begin{definition}
    \label{def:directed_tree_forest_axiom}
    An interpretation $\omega$ is a model of $DirectedTree(R,Root)$ if $\omega_R$\ forms a directed tree whose root (i.e.\ the node with no $R$-parent) is the unique domain element interpreted to be true in $\omega_{Root}$.
\end{definition}

\begin{definition}
    An interpretation $\omega$ is a model of $DirectedForest(R)$ if %$\omega$ is a model of $\Phi$ and
    $\omega_R$\ forms a directed forest.
\end{definition}
 
Using the definition of directed trees, as DAGs with exactly one root and with all the other nodes having indegree 1, we get the following proposition.

\begin{proposition}\label{prop:directed_tree}
    $\Phi \land DirectedTree(R, Root)$ is equivalent to 
    \begin{equation}
        \label{eq: directed_tree}
        \Phi \land Acyclic(R) \land |Root|=1 \land \forall x. \neg Root(x) \rightarrow \exists^{=1} y. R(y,x)
    \end{equation}
     for any sentence $\Phi$. Hence, WFOMC with directed tree axiom can be modularly reduced to WFOMC with DAG axiom.
\end{proposition}

\begin{proof} Let $\omega$ be a model of formula \eqref{eq: directed_tree}, so that $\omega_R$ is a DAG. Since $\omega \models \forall x. \neg Root(x) \rightarrow \exists^{=1} y. R(y,x)$, we have that any element $c$ such that $\omega \models \neg Root(c)$, has exactly one $R$-parent. Since $\omega \models (|Root| = 1)$, we have that $\omega_R$ is a DAG with only one node, say $r$, such that $r$ does not have exactly one $R$-parent, and all other nodes have exactly one  $R$-parent. But $\omega_R$ is a DAG, hence it necessarily has one node with no parent, which hast to be $r$. Hence, $\omega_R$ is a directed rooted tree. It can also be checked that any $\omega \models \Phi$ such that $\omega_R$ is directed rooted tree is a model of formula \eqref{eq: directed_tree}.
\end{proof}

Analogously, using the definition of directed forests, as DAGs where each node has indegree at most 1, we get the following proposition:
\begin{proposition}\label{prop:directed_forest}
    $\Phi \land DirectedForest(R)$ is equivalent to $$\Phi \land Acyclic(R) \land \forall x.\exists^{<=1}y. R(y,x)$$ for any sentence $\Phi$. Hence, WFOMC with directed forest axiom can be modularly reduced to WFOMC with DAG axiom.
\end{proposition}

Since both $DiretedTree(R,Root)$ and $DirectedForest(R)$ can be modularly reduced to WFOMC with DAG axiom and additional C$^2$ constraints, the domain liftability results for the DAG axiom --- such as Theorem \ref{thm: counting_acyclic} --- also transfer to $DiretedTree(R,Root)$ and $DirectedForest(R)$ axioms.

\begin{remark}
    In the definition above we used the predicate $Root$ only for directed trees. Notice, however, that we could define it also for directed forests, by imposing the satisfaction of the following sentence $$\forall x. Root(x) \leftrightarrow \neg \exists y. R(y,x)$$ In a similar way, one could define a unary predicate $Leaf$ for representing leaves, both for directed trees and for directed forests, using the following sentence: $$\forall x. Leaf(x) \leftrightarrow \neg \exists y. R(x,y)$$
   
\end{remark}

\subsection{WFOMC with Tree Axiom}
We will now focus on the undirected tree axiom. 

    \begin{definition}
        \label{def:tree_axiom}
        Let $R$ be a binary predicate, an interpretation $\omega$ is a model of
        $Tree(R)$ if 
        \begin{itemize}
            \item $\omega_R$ \/ forms a symmetric and antireflexive relation of $R$, and 
            \item $\omega_R$ \/ forms a Tree 
        \end{itemize}
    \end{definition}

\begin{remark}
    Notice that one could also define a unary predicate $Leaf$ for representing leaves by imposing the satisfaction of the following sentence: $$\forall x. Leaf(x) \leftrightarrow \exists^{=1} y. R(x,y)$$
    This applies also to the case of undirected forests, which will be introduced later.
\end{remark}

Note that a connected graph on $n$ nodes is a tree if and only if it has $n-1$ edges \cite[Theorem 2.1]{GraphTheory}.  This gives us the following proposition:
\begin{proposition}\label{prop:tree}
    For any sentence $\Phi$ on a domain of cardinality $n$, we have that $\Phi \land Tree(R)$ is equivalent to $$\Phi \land Connected(R) \land |R| = 2n-2$$ Hence, WFOMC with tree axiom can be modularly reduced to WFOMC with connectivity axiom.
\end{proposition}

\begin{remark}\label{rem:liftability_tree}
Since $Tree(R)$ can be modularly reduced to WFOMC with connectivity axiom and additional cardinality constraints, the domain liftability results for the connectivity axiom --- such as Theorem \ref{thm: con_c2} --- also transfer to the $Tree(R)$ axiom. These results were already established in \cite{Tree}, but we are providing a different way to arrive at them.
\end{remark}

\subsection{WFOMC with Forest Axiom}
We now present an algorithm for extending domain liftability of C$^2$ with an undirected forest axiom.
We first prove the formula for counting undirected forests as presented in \cite{Counting_Forests}. This will then form the basis of our algorithm for WFOMC with forest axiom, which is subsequently presented.

\begin{proposition}\label{prop:counting_forests}
Let $t_m$ denote the number of trees on $m$ labeled nodes, and let $f_n$ be the number of forests on $n$ labeled nodes. Then, 
\begin{equation}\label{eq:forests}
    f_n = \sum_{m=1}^n \binom{n-1}{m-1} t_m f_{n-m}
\end{equation}
\end{proposition}
\begin{proof}
    Without loss of generality, we can assume the $n$ labeled nodes to be the set $[n]$. We will prove the proposition by observing that any forest on such nodes can be viewed as a tree containing the node $1$ plus a forest made up of the remaining connected components. 
    Consider an arbitrary forest on $[n]$, and let $m$ denote the number of nodes of the connected component containing the node $1$. Clearly, for an arbitrary forest, $m$ can be any number from $1$ to $n$. 
    Now, fixing $m$, the number of different trees containing node $1$ and $m-1$ other nodes is given by $\binom{n-1}{m-1} t_m$, where $\binom{n-1}{m-1}$ represents the number of ways to choose $m-1$ nodes from $[\bar 1]$, and $t_m$ denotes the number of different trees on the $m$ selected nodes ($1$ included).
    On the other hand, there are $f_{n-m}$ different forests on the remaining $n-m$ nodes, independently of the tree containing $1$. 
    It follows that the number of forests such that the connected component containing $1$ has $m-1$ other nodes is $\binom{n-1}{m-1} t_m f_{n-m}$. To count all possible forests on $n$ nodes, we only need to sum up the contributions from each $m$, and this gives us the desired recursive formula.
\end{proof}

Note that the number $t_m$ of trees on $m$ nodes is given by $m^{m-2}$ (Cayley's formula \cite{Counting_Forests}). However, this is irrelevant for what follows.

\begin{definition}
    \label{def:forest_axiom}
    Let $R$ be a binary predicate. An interpretation $\omega$ is a model of $Forest(R)$ if $\omega_R$ forms a forest. %For any subset $C\subseteq [n]$, an interpretation $\omega$ is a model of $Tree(R, C)$ if $\omega$ is a model of $Forest(R)$ on $[n]$ and $\omega_R \downarrow C$ forms a connected component of $\omega_R$ (and hence a tree).
\end{definition}

\begin{remark}\label{rem:symmetric_forest}
    As done before for connectivity, since $\forall xy. \Phi(x,y) \land Forest(R)$ assumes that $\omega_R$ forms a symmetric and anti-reflexive relation, we can assume without loss of generality that
    \begin{equation}\label{eq:symmetric_forest}
        \forall xy. \Phi(x,y) \models \forall x. \neg R(x,x) \land \forall xy. R(x,y) \rightarrow R(y,x)
    \end{equation}
\end{remark}

\begin{definition}
    \label{def: atleast_m_forest}
    Let $\Psi \coloneqq \forall xy.\Phi(x,y) \land Forest(R)$, where $\Phi(x,y)$ is a quantifier-free FO$^2$ formula interpreted on the domain $[n]$, such that \eqref{eq:symmetric_forest} holds. For any $m \leq n$, we say that an interpretation $\omega$ is a model of\, $\Psi_{[m]}$, and write $\omega \models \Psi_{[m]}$, if  $\omega$ is a model of $\Psi$ on $[n]$ and $\omega_R \downarrow [m]$ forms a connected component of $\omega_R$ (and hence a tree).
\end{definition}

It is easy to check that $\Psi_{[m]}$ in Definition \ref{def: atleast_m_forest} is the same as $\Psi_{[m]}$ in Lemma \ref{lem: count_split_wmc}, where we instantiate  $axiom'$, $axiom''$ and $\Theta(x,y)$ as follows:
    \begin{itemize}
        \item $axiom' \coloneqq Tree(R)$
        \item $axiom'' \coloneqq Forest(R)$
        \item $\Theta(x,y) \coloneqq \neg R(x,y)$
    \end{itemize}
A formal proof can be found in the appendix, as Lemma \ref{lem: atleast_m_forest}.
Using this characterization and Lemma \ref{lem: count_split_wmc}, we have the following proposition.

\begin{proposition}\label{prop: forest}
Let $\Psi \coloneqq \forall xy.\Phi(x, y) \land Forest(R)$, and $\Psi' \coloneqq \forall xy.\Phi(x, y) \land Tree(R)$, where $\Phi(x,y)$ is a quantifier free FO$^2$ formula  satisfying the condition in Remark \ref{rem:symmetric_forest}.
Then:
\begin{equation}\label{eq:Psi_m_forest}
    \wfomc(\Psi_{[m]}, \bk) = \sum_{\substack{\bk=\bk' + \bk'' \\ |\bk'|=m}} \prod_{i,j} r^{\bk'_i \bk''_j}_{ij} \wfomc(\Psi', \bk') \wfomc(\Psi, \bk'') 
\end{equation}
where 
$r_{ij} := \sum_{l} n_{ijl}v_l$, and $n_{ijl}$ is $1$ if $ijl(x, y) \models \Phi(\{x, y\}) \land \neg R(x, y)$ and $0$ otherwise.
\end{proposition}

\begin{proposition}
Let $\Psi \coloneqq \forall xy.\Phi(x, y) \land Forest(R)$  
be interpreted over $[n]$, where $\Phi(x,y)$ is a quantifier free FO$^2$ formula satisfying the condition in Remark \ref{rem:symmetric_forest}.
Then:
\begin{equation}\label{eq:wfomc_forest}
    \wfomc(\Psi, \bk) = \sum_{m = 1}^n \binom{n-1}{m-1} \wfomc(\Psi_{[m]}, \bk)
\end{equation}
\end{proposition}
\begin{proof}
    The proof idea is essentially the same as the one for proposition \ref{prop:counting_forests}. Let $\Psi_C$ be the generalization of $\Psi_{[m]}$ to a generic subset $C\subseteq [n]$. Clearly, $\omega$ is a model of $\Psi$ if and only if $\omega$ is a model of $\Psi_C$ for a subset $C$ containing the node $1$. Notice also that for any $\omega$ there can be only one subset $C$ such that $\omega_R \downarrow C$ is the connected component of $\omega_R$ containing the node $1$. Hence, 
    \begin{equation}
        \wfomc(\Psi, \bk) = \sum_{\substack{C \subseteq [n] \\ 1\in C}} \wfomc(\Psi_{C}, \bk)
    \end{equation}

    Now, $\wfomc(\Psi_{[m]}, \bk)$ is the same as $\wfomc(\Psi_{C}, \bk)$ for any $C \subseteq [n]$ with $|C| = m$. Furthermore, there are $\binom{n-1}{m-1}$ ways of choosing a subset $C$ of $[n]$ in such a way that it contains $1$. This gives us equation \eqref{eq:wfomc_forest}.
\end{proof}

We provide the pseudocode for computing $\wfomc(\Psi,\bk)$ as Algorithm \ref{alg:algorithm_Forest} in the appendix. The only changes w.r.t Algorithm \ref{alg:algorithm_FO_DAG} are that line 6 is substituted with
\begin{algorithmic}[1]
\setcounter{ALG@line}{5}
    \State ${A[\bp] \gets \!\sum_{m=1}^{|\bp|}\binom{|\bp|-1}{m-1}{\overline{\wfomc}}(\!\Psi_{[m]}, \bp)}$
\end{algorithmic}
and that $\wfomc(\Psi', \bk')$ in line 13 is using WFOMC with tree axiom, whose liftability is ensured by Remark \ref{rem:liftability_tree}. With this in mind, it can be proved that the algorithm runs in polynomial time w.r.t domain cardinality in the same way of Algorithm \ref{alg:algorithm_FO_DAG}. Thus, we can derive the domain liftability of C$^2$ with the forest axiom by using the modular reductions from \cite{broeck2013} and \cite{kuzelka2020weighted} (in the same way as we did for DAG and connected axioms).
\begin{theorem} Let $\Psi\coloneqq \Phi \land Forest(R)$, where $\Phi$ is a C$^2$ formula. Then $\wfomc(\Psi,n)$ can be computed in polynomial time with respect to the domain cardinality.
\end{theorem}

Again, we provide an example to show that our results can be useful in enumerative combinatorics.

\begin{example}\label{ex:forest_without_isolated}
    The number of forests without isolated vertices on $n$ labeled nodes can be computed as
    \begin{equation}
        \fomc(\forall x. \exists y. R(x,y) \land Forest(R), n)
    \end{equation}
    for a FOL language made only of the binary relation $R$. Using the procedure in \cite{broeck2013}, this is reduced to a form to which Algorithm \ref{alg:algorithm_Forest} can be directly applied:
    \begin{equation}\label{eq: forests_without}
        \wfomc(\forall xy. S(x) \lor \neg R(x,y) \land Forest(R), n)
    \end{equation}
    for a FOL language made only of $R/2$ and $S/1$, with $S$ having weights $w(S) = 1$ and $\bar w(S) = -1$, and $R$ having weights $w(R) = \bar w(R) = 1$. In fact, with our implementation, we were able to compute sequence A105784 of OEIS.
\end{example}

%%%%%%%%%%%%%%%%%%%%%%%%%%%%%%%%%%%%%%%%% EXPERIMENTS %%%%%%%%%%%%%%%%%%%%%%%%%%%%%%%%%%%

\section{Experiments}
\label{sec:exp}

In this section we provide an empirical analysis of the practical efficiency and expressivity of the constraints introduced in this paper. We implemented our WFOMC algorithms (Algorithms \ref{alg:algorithm_FO_DAG}, \ref{alg:algorithm_CON}, and \ref{alg:algorithm_Forest}) in Python,\footnote{Code is available at \href{https://github.com/dbizzaro/WFOMC-beyond-FOL}{https://github.com/dbizzaro/WFOMC-beyond-FOL}} and the experiments on an Apple M3 Pro computer. The implementation of the cardinality constraints on unary predicates was done by filtering the $1$-type cardinality vectors $\bk$, as in \cite{AAAI_Sagar}; while for binary predicates we used symbolic weights,\footnote{The use of symbolic weights is equivalent to the procedure in the proof of Theorem \ref{th: cardinality}, as explained in the appendix (to Section 3).} as in \cite{discovering_sequences}.

\subsection{Combinatorial Examples}
\label{subsec:comb}
\begin{figure}
\centering
\includegraphics[width=0.6\textwidth]{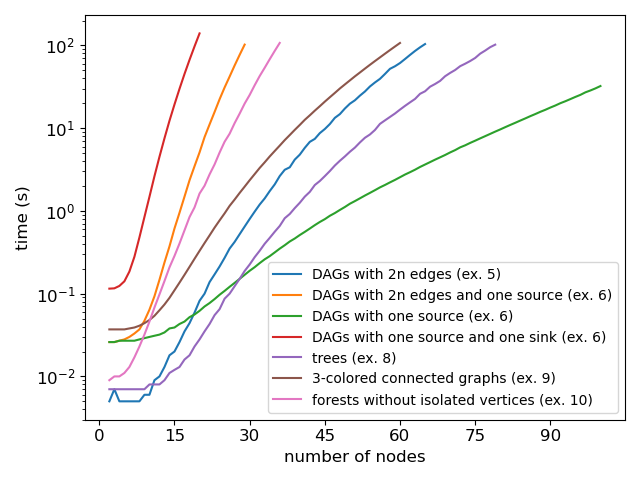}
\caption{Run-times of all experiments as function of the domain cardinality $n$. For the WFOMC encoding see the referred examples. 
The number of active $1$-types for each formula, and if cardinality constraints on binary predicates were used, is reported in Table \ref{tab:active} of the appendix. }
\label{fig: all}
\end{figure}

 We experiment with the combinatorial examples provided in the previous sections, and run-times are reported in Figure \ref{fig: all}. Although our approach provides a general language for easily formulating complex combinatorial problems, and our WFOMC algorithms run in polynomial time w.r.t domain cardinality,  our experiments show that they are far from optimal.
 This is especially pronounced in the run-time for computing the number of trees, which admits the simple closed-form formula $n^{n-2}$. 
In our observation, the number of (active) 1-types\footnote{Active $1$-types for a formula are the 1-types that are consistent with the formula.} and the presence of cardinality constraints on binary predicates constitute key sources of intractability (see Appendix Table \ref{tab:active}). 
As the former leads to more and larger operations,  the latter leads to a complex polynomial interpolation task. Hence, the generality of our proposed framework comes potentially at the cost of lower efficiency, in comparison to results obtained by more fine-grained combinatorics. These results show that more work needs to be done on better practical implementation of the proposed WFOMC results, potentially integrating solutions like \cite{toth2024complexity,faster_lifting}.

\subsection{Markov Logic Networks}
\label{subsec:MLN}
A \emph{Markov Logic Network (MLN)} \cite{richardson2006markov} is a set of weighted formulas $\Phi\coloneqq \{w_i : \phi_i\}_i$, 
where each $\phi_i$ is a quantifier free FOL formula with weight $w_i \in \R \cup \{\infty\}$. The formulas with weight $\infty$ are hard constraints, i.e., any world that does not follow those constraints has probability $0$. Let $\Phi_\R$ be the subset of real-valued weighted formulas in $\Phi$, and let $\Phi_\infty$ be the hard constraints. Given a domain $\Delta$, an MLN $\Phi$ defines a probability distribution over the set of possible interpretations:
\begin{equation}
    P^\Delta_\Phi (\omega) = \frac{\mathbbm{1}_{\omega \models \Phi_\infty}}{Z^\Delta_{\Phi}} \exp\left(\sum_{w_i: \phi_i \in \Phi_\R} w_i \cdot n(\phi_i, \omega)\right)
\end{equation}
where 
\begin{itemize}
    \item $n(\phi_i, \omega)$ represents the number of true groundings of $\phi_i$ in $\omega$;
    \item $\mathbbm{1}_{\omega \models \Phi_\infty}$ is $1$ when $\omega \models \Phi_\infty$ and $0$ otherwise;
    \item $Z^\Delta_\Phi$ is a normalization constant (ensuring that $P^\Delta_\Phi$ is a probability distribution) called \emph{partition function}. 
\end{itemize}

One of the applications of WFOMC is probabilistic inference in MLNs \cite{broeck2013}.  For every finitely-weighted formula $w_i : \phi_i$ in an MLN $\Phi$, we introduce a fresh predicate $P_i/a_i$, whose arity $a_i$ is the number of free variables in $\phi_i$. We define a weight function $(w, \bar w)$ such that $w(P_i) = \exp(w_i)$ and $\bar w(P_i)=1$ for each $i$, while $w(R)=\bar w(R)=1$ for every other predicate $R$. Moreover, let $\Psi$ denote the conjunction of $\Phi_\infty$ with all the formulas

\begin{equation}
    \forall x_1\dots x_{a_i}.\ P_i(x_1, \dots, x_{a_i}) \leftrightarrow \phi(x_1, \dots, x_{a_i}) 
\end{equation}
Then, for any query sentence $\phi$, probabilistic inference can be computed in the following way:
\begin{equation}
    P^\Delta_\Phi (\phi) = \frac{\wfomc(\Psi \land \phi, |\Delta|)}{\wfomc(\Psi, |\Delta|)}
\end{equation}
This means that inference can be performed in polynomial time (w.r.t domain cardinality) whenever $\Psi$ and $\phi$ are domain-liftable.

The constrains introduced in this paper, i.e., $DAG(R)$, $Connected(R)$ and $Forest(R)$, can also be used as hard constraints ($\Psi$) or as queries ($\phi$), while admitting polynomial time inference in MLNs.  The experiments below demonstrate that the probability distributions that can be obtained by incorporating these constraints in an MLN can be difficult to achieve otherwise. Such distributions may result in a better model whenever we have reasons to believe that the graph formed by a particular binary relation should be a DAG, a connected graph or a forest. All the distributions in the experiments below are computed (exactly) using the algorithms introduced in the paper.

\subsection{Graph Statistics}
\begin{figure}[ht]
      \centering
      \subfloat{\includegraphics[width=.48\textwidth]{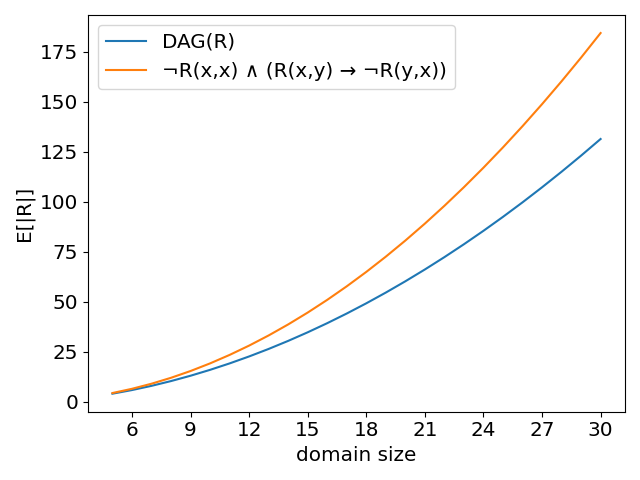}}
      \
      \subfloat{\includegraphics[width=.48\textwidth]{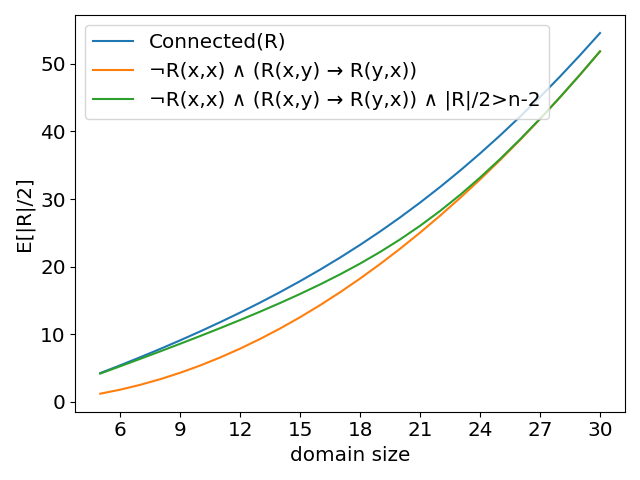}}
      \
      \subfloat{\includegraphics[width=.48\textwidth]{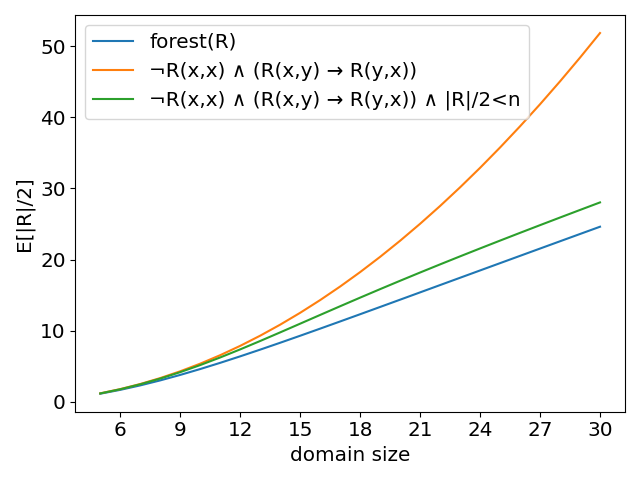}}
  \caption{Expected values of the number of edges of the directed/undirected graphs produced by MLNs with only the predicate $R/2$. The legends report the hard constraints, while the soft constraint is always $-1: R(x,y)$.}
  \label{fig:edges-expectations}
\end{figure}

Let us begin by examining the simplest scenario: an MLN that models only a directed (resp.\ undirected) graph. When restricting the MLN to consider only graphs that are DAGs (resp.\ connected, or forests), then all the statistics about the graphs can be different. An example of this is shown in Figure~\ref{fig:edges-expectations}. In this experiment, we show that MLNs with our constraints lead to different edge distributions compared to the unconstrained cases. 
For this analysis, we considered a fixed prior on the distribution of edges given by the weighted formula $-1:R(x,y)$ (which encodes a preference for sparse graphs), and plotted the distributions of the number of edges. We compared the distributions when the graph constraints are imposed (blue curves), against what are arguably their best approximations in FO$^2$ (orange curves). For connectivity and forests, we compared also the distribution produced by the corresponding FO$^2$ formula together with the cardinality constraint that matches the support of the number of edges for the two cases\footnote{Connected graphs can have any number of edges between $n-1$ and $\binom{n}{2}$, while forests can have any number of edges between $0$ and $n-1$.} (green curves). 
The actual distributions of the edge count for domain size $n=20$ is available in the appendix (Figure \ref{fig:edges}). As expected, the DAG and forest constraints favor more sparsity, while the opposite is true for connectivity. The next example will show that even when this particular effect is not very pronounced, the overall effect of a global constraint like connectivity can be significant (even compared to cardinality constraints).

\subsection{Smokers \& Friends with Connectivity Constraint}

\noindent
We will now analyze the widely investigated social network example of ``smokers and friends'' in the sparse regime. The example is encoded with the following soft constraints: 
\begin{align}
    w_S &: S(x) \\
    w_F &: F(x,y) \\
    w_P &: S(x) \land F(x,y) \to S(y)
\end{align}

where $S/1$ is the predicate for smokers and $F/2$ for friendship.
If the weight $w_P$ is positive, then the MLN models the fact that friends of smokers are more likely to be smokers.
The network can be made sparse by setting the weight $w_{F}$ to be negative --- larger negative weight leads to more sparsity. 

\begin{figure}[ht]
      \centering
      \subfloat{\includegraphics[width=.48\textwidth]{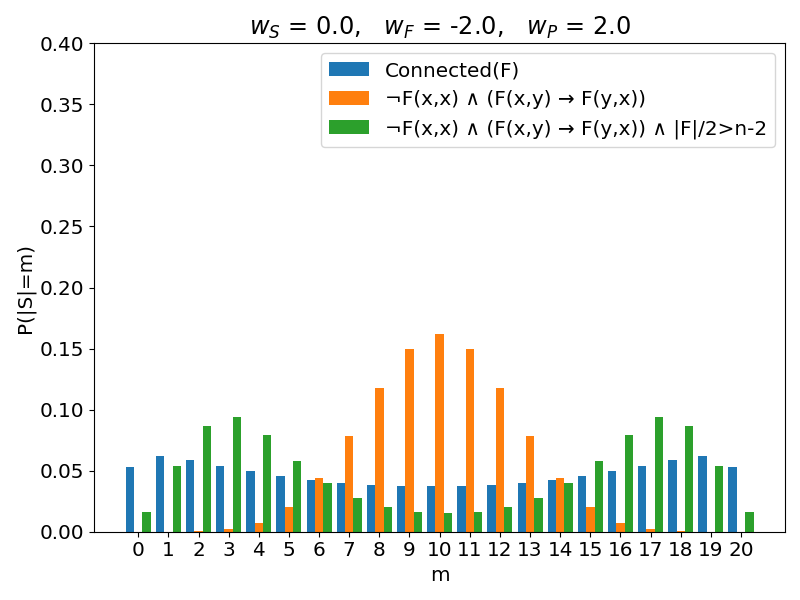}} \
      \subfloat{\includegraphics[width=.48\textwidth]{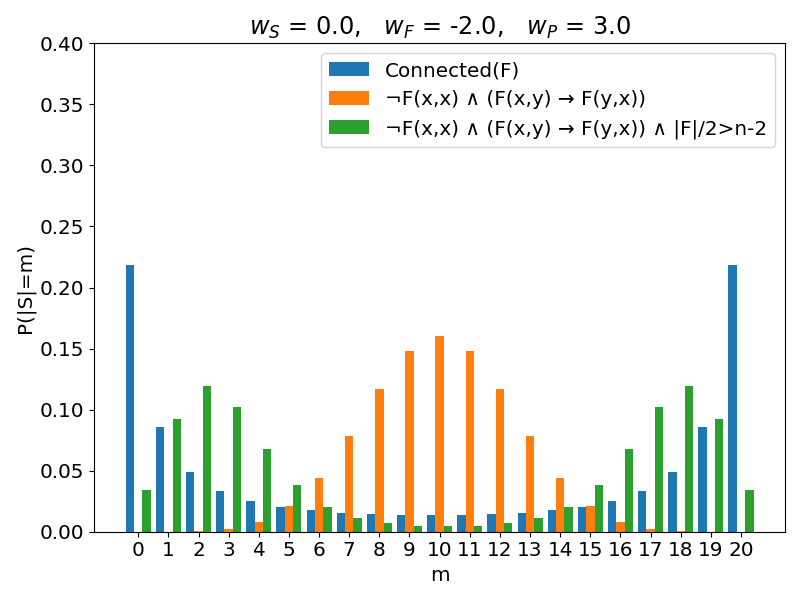}} \
      \subfloat{\includegraphics[width=.48\textwidth]{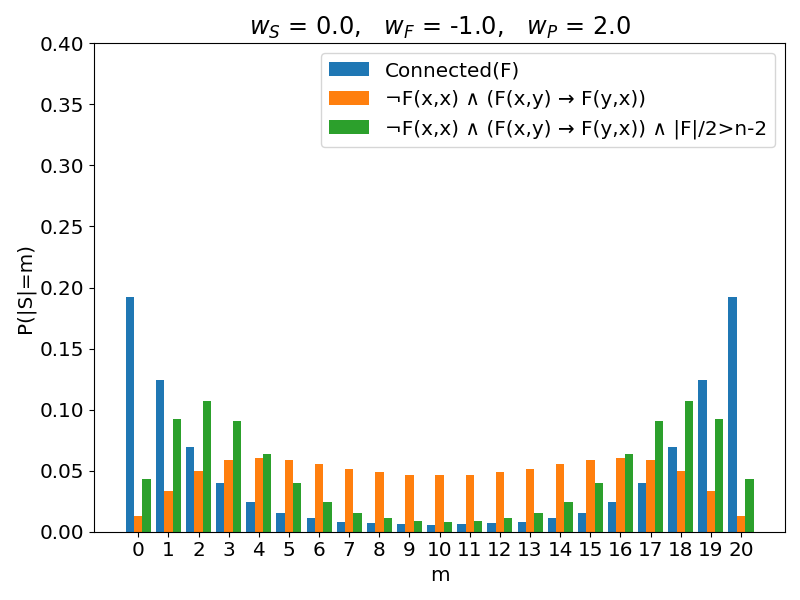}} \
      \subfloat{\includegraphics[width=.48\textwidth]{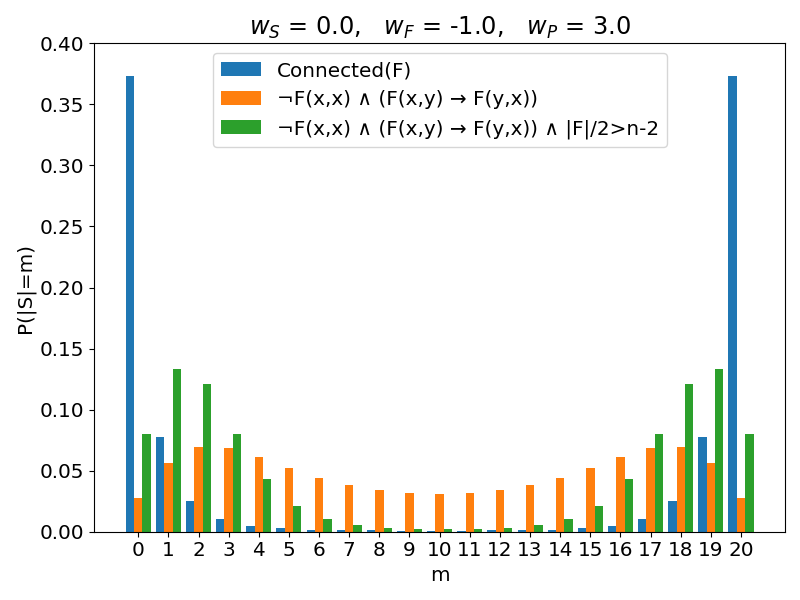}}
  \caption{Distributions of smokers for different weights. The legends report the hard constraints, while the titles report the weights of the soft constraints. The domain size is fixed to $n=20$.}
  \label{fig:smokers}
\end{figure}

The partition function for this MLN can be computed (in polynomial time) as the WFOMC of the following formula:
\begin{equation}\label{eq:hard}
\forall xy.\ P(x,y) \leftrightarrow (S(x) \land F(x,y) \to S(y))
\end{equation}
with the following weight function:
\begin{align}
    w(S) = \exp(w_S), &\quad \bar w(S) = 1 \\
    w(F) = \exp(w_F), &\quad \bar w(F) = 1 \\
    w(P) = \exp(w_P), &\quad \bar w(P) = 1
\end{align}
Figure~\ref{fig:smokers} compares the distributions of the number of smokers for the following three cases\footnote{Each of the three cases is modeled by adding the corresponding hard constraint to the WFOMC of equation \eqref{eq:hard}. The hard constraints are formally expressed in the legends of Figure \ref{fig:smokers}.}:
\begin{itemize}
    \item the graph realized by $F$ must be a connected undirected graph (blue);
    \item the graph realized by $F$ can be any undirected graph (orange);
    \item the graph realized by $F$ must be an undirected graph with at least as many edges as the number of nodes minus one (which is the minimum number of edges for connected graphs; green).
\end{itemize}

In this experiment (Figure~\ref{fig:smokers}), we analyze four different values of weight combinations for $w_F$ and $w_P$, while $w_S$ is fixed to be zero. 
We can see that the distribution becomes more concentrated at the extremes (either all smokers or none) when: ($i$) increasing the bias towards the satisfaction of the ``smoker and friends'' formula (i.e.\ increasing $w_P$), or ($ii$) decreasing the sparsity of the friendship network (i.e.\ increasing $w_F$). This aligns with our intuition: ($i$) as $w_P$ increases, any connected component of the friendship network is more likely to have either all smokers or none; ($ii$) the less sparsity (i.e., the larger $w_{F}$), the more likely to have one large connected component comprising most of the nodes --- and all the nodes of such component tend to become homogeneously smokers or non-smokers, modulated by the weight $w_P$. The interplay of $w_F$ and $w_P$ controls the connectivity and smoking-homogeneity of the networks modeled by the MLN.

Given that connectivity controls the \say{spreading} of homogeneity of smoking, adding connectivity constraints allow us to further modulate the global distribution of smokers in the network --- as reflected in the blue histograms in Figure~\ref{fig:smokers}. 
Note that the connected networks still maintain high sparsity, as evidenced in Figure~\ref{fig:edges-expectations}, and shown more precisely in the appendix (Figure~\ref{fig:edges-smokers}, top-right). 
Importantly, connectivity is forcing the graph to have only one connected component, so smoking  behavior can \say{spread} across the entire graph, making most of the nodes either smokers or non-smokers. 
As expected, the distribution of smokers under the connectivity constraint is consistently more concentrated at the extremes than in the other two cases.

Analogous experiments with DAG and forest constraints instead of connectivity are reported in the appendix (Figures \ref{fig:smokers-DAG} and \ref{fig:smokers-forest}). In these cases, the differences are mainly driven by the corresponding increase in sparsity. Although these constraints demonstrate different artifacts than FOL and connectivity, we leave there qualitative analysis to future work for more relevant examples. For instance, citation networks for DAGs, and genealogy networks for forests.

\begin{figure}[ht]
\centering
\includegraphics[width=0.48\textwidth]{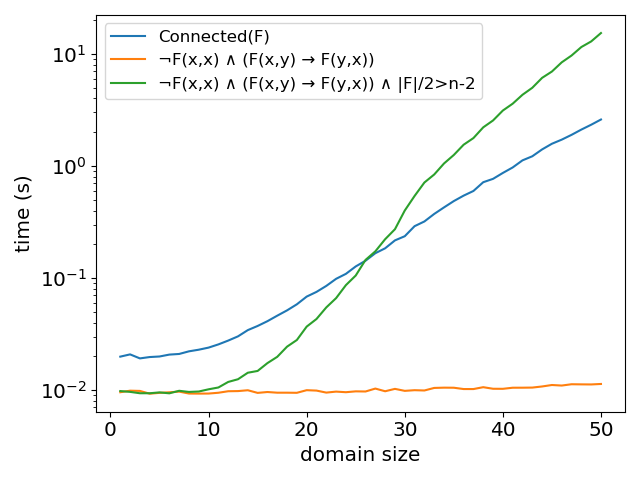}
\caption{Run-times of computing the partition function of the ``smokers and friends'' MLNs.}
\label{fig: runtimes_smokers}
\end{figure}

Finally, Figure~\ref{fig: runtimes_smokers} shows the run-times of computing the partition function of the ``smokers and friends'' MLNs, i.e.\ the run-times of the WFOMC of each of the expressions in the legend, together with formula \eqref{eq:hard}. To avoid numerical issues, the weights were represented as fractions, and exact calculations were performed.

\section*{Conclusion}
We investigate the domain-liftability of the first order logic fragment with two variables and counting quantifiers (C$^2$), with additional graph theoretical constraints. We show that the domain liftability of C$^2$ is preserved when one of the relations in the language is restricted to represent an acyclic graph, a connected graph, a forest (resp. a directed forest), or a tree (resp. a directed tree). A key novel idea used consistently throughout our work is \emph{counting by splitting}. The generality of this principle can potentially aid the lifting of many other novel constraints. Besides their application in statistical relational learning, our results  provide a uniform framework for combinatorics on different types of multi-relational graphs. In fact, our work covers and extends a vast array of results in combinatorics, such as counting phylogenetic networks and enumerating acyclic graphs/forests with different constraints. Our work (along with \cite{Tree}) can be seen as extending the relational language for constraints that admit lifted inference beyond first order logic. Hence, these results motivate the following open problem:
\begin{center}
\emph{Is there a formal language that captures all tractable counting problems for labeled relational structures?}
\end{center}

Our experimental analysis shows that the new tractable constrains lead to potentially useful differences in distributions expressed by Markov Logic Networks.
However, we observe that our algorithms, though polynomial in time, have scalability issues, and often do not lead to the optimal computational complexity. This motivates further research into designing more efficient practical implementations, and in understanding the gap between domain lifted and optimal inference. 

\bibliographystyle{elsarticle-num}
\bibliography{elsarticle-template-num}

%%%%%%%%%%%%%%%%%%%%%%%%%%%% APPENDIX %%%%%%%%%%%%%%%%%%%%%%%%%%%%%%%%

\section*{Appendix}

\section*{Appendix to Section \ref{sec: background}}

\thmCardinality*
\begin{proof} 
\label{proof: cardinality}
Let us consider an FOL language $\mathcal{L}$ that contains $r$ relational symbols denoted by $\{R_i/a_i\}_{i \in [r]}$. Let $\omega$ be an interpretation and let $\bm{\mu} = \langle |R_1|, \dots, |R_{r}|\rangle$ be the vector comprising the cardinality of each predicate $R_i$ in $\omega$. Now, $\w(\omega)$ can be evaluated using the definition of symmetric weight functions (Definition \ref{def: symm}). Any two interpretations that have the same predicate cardinalities $\bm{\mu}$ as $\omega$ have the same weight $\w(\omega)$. Therefore, we use $\w_{\bm{\mu}}$ to indicate the weight $\w(\omega)$, where $\omega \models \bm{\mu}$. Given an FOL formula $\Phi$, let $A_{\bm{\mu}}$ be the number of interpretations $\omega \models \Phi \land \bm{\mu}$, then  the following holds:
\begin{equation}
    \label{eq: cardinality}
    \wfomc(\Phi,\bk) = \sum_{\bm{\mu}}A_{\mu}\w_{\bm{\mu}}
\end{equation}

For each predicate $R_i/a_i$ in the FOL language $\mathcal{L}$, there exist $n^{a_i}$ ground atoms. Therefore, there are $n^{\sum_{i\in[r]}a_i}$ potential values of $\bm{\mu}$, which means that there are polynomially many vectors $\bm{\mu}$ with respect to $n$. By evaluating $\wfomc(\Phi,\bk)$ for $n^{\sum_{i\in[r]}a_i}$ distinct weight function pairs $(w,\bar{w})$, we can obtain a non-singular linear system of $n^{\sum_{i\in[r]}a_i}$ equations on the $n^{\sum_{i\in[r]}a_i}$ variables $A_{\bm{\mu}}$. This system can be solved using Gauss-elimination algorithm in $O(n^{3\sum_{i\in[r]}a_i})$ time. Then, having found the numbers $A_{\bm{\mu}}$, we can compute the value of any cardinality constraint as follows:
\begin{equation}
    \label{eq: cardinality_constraint}
    \wfomc(\Phi \land \Gamma,\bk) = \sum_{\bm{\mu \models \Gamma}}A_{\mu}\w_{\bm{\mu}}
\end{equation}
where $\bm{\mu \models \Gamma}$ represents the fact that the predicate cardinalities $\bm{\mu}$ satisfy the cardinality constraint $\Gamma$. Since, there is only a polynomial number of vectors $\bm{\mu}$, equation \eqref{eq: cardinality_constraint} can be computed in polynomial time. 
\end{proof}

\begin{remark} In equation \eqref{eq: cardinality_constraint} we assume that $\bm{\mu} \models \Gamma$ can be checked in polynomial time w.r.t.\ $n$. This is a reasonable assumption for all our purposes.  
\end{remark}

\subsection*{Practical Implementation of Cardinality Constraints}

Note that the proof of \Cref{th: cardinality} assumes only a black-box access to the $\wfomc(\Phi,\bk)$ oracle. However, for all the algorithms introduced in the paper we can treat $\wfomc(\Phi,\bk)$ as a polynomial with symbolic weights for each predicate as variables.  For example, let $\Phi \coloneqq \forall xy.  \neg R(x,x) \land (R(x,y) \rightarrow R(y,x))$ (encoding for an undirected graph without self-loops). Then, equation \eqref{eq: WFOMC_Beame} becomes:
\begin{equation*}
\label{eq: cardinality_example}
    \wfomc (\Phi, (w,\bar{w}), n) = (w(R)^2 + \bar{w}(R)^2)^{\binom{n}{2}} = \sum_{k=1}^{\binom{n}{2}}\binom{\binom{n}{2}}{k}w(R)^{2k} \bar{w}(R)^{2(\binom{n}{2}-k)}
\end{equation*}
which is a polynomial in the variables $w(R)$ and $\bar w(R)$.
Similarly to equation~\eqref{eq: cardinality}, we have that this polynomial can be written as $\sum_{\bm{\mu}}A_{\bm\mu}\w_{\bm{\mu}}$.
The monomial $\binom{\binom{n}{2}}{k}w(R)^{2k} \bar{w}(R)^{2(\binom{n}{2}-k)}$ is  equal to $A_{\bm\mu}\w_{\bm\mu}$, where $\bm\mu$ is such that $|R| = 2k$, $\w_{\bm\mu} = w(R)^{2k} \bar{w}(R)^{2(\binom{n}{2}-k)}$ and hence $A_{\bm\mu} = \binom{\binom{n}{2}}{k}$. This procedure of inferring $\bm{\mu}$ from monomials is trivially implied from \Cref{thm: beam}. All WFOMC algorithms presented in this paper and in literature \cite{kuzelka2020weighted, AIxIA, AAAI_Sagar} only add simple arithmetic operations over the polynomials obtained in \Cref{thm: beam}. Hence, inferring $\bm{\mu}$ from monomials in the WFOMC polynomials is always possible. Keeping only the monomials satisfying the condition $\bm{\mu} \models \Gamma$ can then be used to impose any cardinality constraint $\Gamma$ as done in \Cref{eq: cardinality_constraint}. For instance, for a cardinality constraint $\Gamma$, one can simply eliminate the terms {$\binom{\binom{n}{2}}{k}w(R)^{2k} \bar{w}(R)^{2(\binom{n}{2}-k)} = A_{\bm\mu}\w_{\bm{\mu}}$} such that $\bm{\mu} \not\models \Gamma$. Leaving us with $\sum_{\bm{\mu} \models \Gamma }A_{\bm\mu}\w_{\bm{\mu}}$.

In general, one can always consider the predicate weights as variables, and the WFOMC algorithms will produce a polynomial on such weights. Once simplified, this polynomial can be expressed as in the RHS of equation \eqref{eq: cardinality}, with the difference that now the coefficients $A_{\bm{\mu}}$ are known (since they are produced by the WFOMC algorithms and the polynomial simplification), while the weights $\w_{\bm{\mu}}$ are symbolic. Thus, in our implementation, instead of computing the coefficients $A_{\bm{\mu}}$ as in the proof of \Cref{th: cardinality}, we proceed in the following manner:
with sympy Python library, we represent symbolically the predicate weights, and then perform the WFOMC operations as operations between polynomials on such variables. This produces a polynomial that when simplified gives us directly the coefficients $A_{\bm{\mu}}$ and hence a simple way to impose any cardinality constraint as done in \Cref{eq: cardinality_constraint}.

\section*{Appendix to Section \ref{sec: DAG}}

\begin{lemma}
    \label{lem: atleast_m_DAG}
    An interpretation $\omega$ is a model of $\Psi_{[m]}$ according to Definition \ref{def: atleast_m_shorter} if and only if $\omega$ satisfies conditions C1-C4 in  Lemma \ref{lem: count_split_wmc}, where we instantiate 
    $axiom'$, $axiom''$ and $\Theta(x,y)$ as follows:
    \begin{itemize}
        \item $axiom' \coloneqq \forall xy. \neg R(x,y)$
        \item $axiom'' \coloneqq Acyclic(R)$
        \item $\Theta(x,y) \coloneqq \neg R(y,x)$
    \end{itemize} 
\end{lemma}
\begin{proof} 
If $\omega \models \Psi_{[m]}$ according to Definition \ref{def: atleast_m_shorter}, since nodes in $[m]$ have zero indegree, we have that $\omega \downarrow [m] \models axiom'$ (C1). Since any subgraph of an acyclic graph is acyclic, we also have that $\omega \downarrow [\bar{m}] \models axiom''$ (C2). And since nodes in $[m]$ have indegree zero, there can be no arrow from $[\bar{m}]$ to $[m]$, i.e., $\omega \models \forall x \in [m]\,\, \forall y \in [\bar m]. \Theta(x,y)$ (C3). Finally, C4 is satisfied by construction. On the other hand, if $\omega$ satisfies C1 and C3 with the previous instantiations, then the nodes in $[m]$ cannot have incoming arrows. And the graph obtained from a DAG by adding $m$ nodes without incoming arrows is still a DAG. 
\end{proof}

\section*{Appendix to Section \ref{sec: connected}}
\begin{lemma}
    \label{lem: atleast_m_connected}
    An interpretation $\omega$ is a model of $\Psi_{[m]}$ according to Definition \ref{def: atleast_m_connected} if and only if $\omega$ satisfies conditions C1-C4 in  Lemma \ref{lem: count_split_wmc}, where we instantiate 
    $axiom'$, $axiom''$ and $\Theta(x,y)$ as follows:
    \begin{itemize}
        \item $axiom' \coloneqq Connected(R)$
        \item $axiom'' \coloneqq \top$
        \item $\Theta(x,y) \coloneqq \neg R(x,y)$
    \end{itemize} 
\end{lemma}
\begin{proof} If $\omega \models \Psi_{[m]}$ according to Definition \ref{def: atleast_m_connected}, since $[m]$ is a connected component, we have that $\omega \downarrow [m] \models axiom'$. Also, $axiom''$ is vacuously satisfied.   Since $[m]$ is a connected component of $\omega_R$, there can not be an $R$-edge between $[m]$ and $[\bar{m}]$ -- as that would make $[m]$ part of a larger connected subgraph, leading to a contradiction. Hence, $\omega \models \forall x \in [m]\,\, \forall y \in [\bar m]. \Theta(x,y)$. Finally, C4 is satisfied by construction. On the other hand, if $\omega$ satisfies C4, then it satisfies the condition in Remark \ref{rem: symmetric_con} for $R$ to be a symmetric and antireflexive relation. So, by satisfying also C1 and C3, the nodes in $[m]$ must be connected and there cannot be edges between them and the other $n-m$ nodes. This means that precisely that $\omega_R \downarrow [m]$ forms a connected component of $\omega_R$.
\end{proof}

\subsection*{Analysis of the Algorithm and Proof of Theorem \ref{thm: con_c2}}

We provide Algorithm \ref{alg:algorithm_CON} that takes as input $\Psi\coloneqq \forall xy. \Phi(x,y) \land Connected(R)$ and $\bk$ -- where $\Phi(x,y)$ is a quantifier-free FO$^2$ formula and $\bk$ is a 1-type cardinality vector with $|\bk| = n$ -- and returns $\wfomc(\Psi,\bk)$. It can be seen that  the algorithm runs in polynomial time using much of the same analysis that was used for Algorithm~\ref{alg:algorithm_FO_DAG}. The key ideas being that: the $\mathbf{for}$ loops in line 5-8 and line 12-14 run both polynomially many iterations w.r.t $n$; the lexicographical order in  the $\mathbf{for}$ loop on line 5 ensures that the values $A[\bs']$ required in the function $\overline{\wfomc}(\Psi_{[m]},\bs)$ are always already stored in $A$; and $\wfomc(\forall xy. \Phi(x,y),\bs'')$ is an FO$^2$ WFOMC problem, again computable in polynomial time. 

\begin{algorithm}[htb]
    \caption{WFOMC-Connected}
    \label{alg:algorithm_CON}
    \begin{algorithmic}[1]
    \State \textbf{Input}: $\Psi\coloneqq \forall xy. \Phi(x,y) \land Connected(R), \bk$
    \State \textbf{Output}: $\wfomc(\Psi,\bk)$
        \State $A[\mathbf{0}] \gets 0$ \Comment{$A$ has $u$ indices}
        \State \Comment{$\mathbf{0} = \langle 0,...,0 \rangle$ }
    
        \For{$\mathbf{0} < \bp \leq \bk$ where $ \bp \in \mathbb{N}_{0}^{u}$} \Comment{$\leq$ is Lexical order} 
        \State ${\!A[\bp] \gets \wfomc(\Psi'', \bp) - \frac{1}{|\bp|}\sum_{m=1}^{|\bp|-1}\binom{|\bp|}{m}\cdot m \cdot {\overline{\wfomc}}(\!\Psi_{[m]},\bp)}$ 
        \EndFor\\
        \textbf{return} $A[\bk]$ 
        \State
        \Function{$\overline{\wfomc}$}{$\Psi_{[m]}$, $\bs$} \Comment{Equation \eqref{eq: con_k_k'_m}}
        \State $S = 0$
        \For{$\bs' + \bs'' = \bs$ and $|\bs'| = m$}
        \State $S \gets S + A[\bs'] \cdot \wfomc(\Psi'',\bs'') \cdot \prod_{i,j\in[u]} r_{ij}^{s'_is''_j}$   
        \EndFor
        \State \Return $S$
        \EndFunction
        \end{algorithmic}
\end{algorithm}

In line $3$, an array $A$ with $u$ indices is initiated and $A[\mathbf{0}]$ is assigned the value  $0$, where $\mathbf{0}$ corresponds to the $u$-dimensional zero vector. The for loop in lines $5-7$ incrementally computes $\wfomc(\Psi,\bp)$, using equation \eqref{eq: WFOMC_CON_final}, where the loop runs over all $u$-dimensional integer vectors $\bp$, such that $\bp \leq \bk$, where $\leq$ is the lexicographical order. The number of possible $\bp$ vectors is at most $n^{u}$. Hence, the  for loop runs at most $n^{u}$ iterations. In line 6, we compute $\wfomc(\Psi, \bp)$ as given in equation \eqref{eq: WFOMC_CON_final}. Also in line 6,  the function $\overline{\wfomc}(\Psi_{[m]},\bp)$ --- that computes $\wfomc(\Psi_{[m]},\bp)$ ---is called at most $|\bp|-1$ times, which is bounded above by $n$. $A[\bp]$ stores the value $\wfomc(\Psi,\bp)$.  In the function $\overline{\wfomc}(\Psi_{[m]},\bs)$, the number of iterations in the for loop is bounded above by $n^{2u}$. And $\wfomc(\Phi,\bs'')$ is an FO$^2$ WFOMC problem, again computable in polynomial time. Hence, the algorithm WFOMC-Connected runs in polynomial time w.r.t domain cardinality. Notice that since loop 5-7 runs in lexicographical order, the $A[\bs']$ required in the function $\overline{\wfomc}(\Psi_{[m]},\bs)$ are always already stored in $A$. 

Summing $\wfomc(\Psi,\bk)$ over all possible $\bk$ such that $|\bk| = n$, we can compute $\wfomc(\Psi, n)$ in polynomial time w.r.t domain cardinality $n$. Moreover, due to the modularity of the skolemization process for WFOMC \cite{broeck2013},  we can extend this result to prove the domain liftability of the entire FO$^2$ fragment, with connectivity axiom. 
Using Theorem \ref{th: cardinality} and Remark \ref{rem: FOL_inexpressible_Card}, we can also extend domain-liftability of FO$^2$, with connectivity axiom and cardinality constraints.
Finally, since WFOMC of any C$^2$ formula can be modularly reduced to WFOMC of an FO$^2$ formula with cardinality constraints \cite{kuzelka2020weighted}, we have Theorem \ref{thm: con_c2}.

\section*{Appendix to Section \ref{sec: forest}}

\begin{lemma}
    \label{lem: atleast_m_forest}
    An interpretation $\omega$ is a model of $\Psi_{[m]}$ according to Definition \ref{def: atleast_m_forest} if and only if $\omega$ satisfies conditions C1-C4 in  Lemma \ref{lem: count_split_wmc}, where we instantiate 
    $axiom'$, $axiom''$ and $\Theta(x,y)$ as follows:
    \begin{itemize}
        \item $axiom' \coloneqq Tree(R)$
        \item $axiom'' \coloneqq Forest(R)$
        \item $\Theta(x,y) \coloneqq \neg R(x,y)$
    \end{itemize} 
\end{lemma}
\begin{proof} If $\omega \models \Psi_{[m]}$ according to Definition \ref{def: atleast_m_forest}, since $\omega_R \downarrow [m]$ is a tree, we have that $\omega \downarrow [m] \models axiom'$. Also, any subgraph of a forest is still a forest, so  $\omega \downarrow [\bar m] \models axiom''$.   Since $[m]$ is a connected component of $\omega_R$, there can not be an $R$-edge between $[m]$ and $[\bar{m}]$. Hence, $\omega \models \forall x \in [m]\,\, \forall y \in [\bar m]. \Theta(x,y)$. Finally, C4 is satisfied by construction. On the other hand, if $\omega$ satisfies C4, then it satisfies the condition in Remark \ref{rem:symmetric_forest} for $R$ to be a symmetric and antireflexive relation. Hence, by satisfying also C1, C2 and C3, the nodes in $[m]$ form a tree (C1), those in $[\bar m]$ form a forest (C2), and the two subgraphs are disconnected (C3). This implies that $\omega \models \Psi_{[m]}$ according to Definition \ref{def: atleast_m_forest}.
\end{proof}

\begin{algorithm}[H]
    \caption{WFOMC-Forest}
    \label{alg:algorithm_Forest}
    \begin{algorithmic}[1]
    \State \textbf{Input}: $\Psi\coloneqq \forall xy.\Phi(x, y) \land Forest(R), \bk$
    \State \textbf{Output}: $\wfomc(\Psi,\bk)$
        %\State
        \State $A[\mathbf{0}] \gets 1$ \Comment{$A$ has $u$ indices}
        \State \Comment{$\mathbf{0} = \langle 0,...,0 \rangle$ }
        \For{$\mathbf{0} < \bp \leq \bk$ where $ \bp \in \mathbb{N}_{0}^{u}$} \Comment{Lexical order} 
        \State ${\!A[\bp] \gets \!\sum_{m=1}^{|\bp|}\binom{|\bp|-1}{m-1}{\overline{\wfomc}}(\!\Psi_{[m]}, \bp)}$ 
        \EndFor\\
        \textbf{return} $A[\bk]$ 
        \State
        \Function{$\overline{\wfomc}$}{$\Psi_{[m]}$, $\bs$} \Comment{Equation \eqref{eq:Psi_m_forest}}
        \State $S = 0$
        \For{$\bs' + \bs'' = \bs$ and $|\bs'| = m$}
        \State $S \gets S + \wfomc(\Psi',\bs')\cdot A[\bs''] \cdot \prod_{i,j\in[u]} r_{ij}^{s'_is''_j} $ 
        \EndFor
        \State \Return $S$
        \EndFunction
        \end{algorithmic}
\end{algorithm}

\section*{Appendix to Section \ref{sec:exp}}

\begin{table}[ht]
    \centering
    \begin{tabular}{c|c|c|c}
    Combinatorial problem & \# $1$-types & cardinality $R/2$ & max $n$ \\
    \hline
    
        DAGs w/ $2n$ edges \eqref{eq: DAGs_edges} & $1$ & yes & $65$\\
        DAGs w/ $2n$ edges and one source  \eqref{eq: DAGs_edges_one_source} & $3$ & yes & $29$\\
        DAGs w/ one source  \eqref{eq: DAGs_one_source} & $3$ & no & $122$ \\
        DAGs w/ one source and one sink  \eqref{eq: DAGs_one_source_one_sink} & $9$ & no & $20$\\
        trees  \eqref{eq: connected_tree} & $1$ & yes & $79$\\
        3-colored connected graphs  \eqref{eq: 3_colored} & $3$ & no & $60$\\
        forests w/o isolated nodes  \eqref{eq: forests_without} & $2$ & yes & $36$
    \end{tabular}
    \caption{Number of active $1$-types (\# $1$-types) and presence of cardinality constrains on binary predicates (cardinality $R/2$) for each combinatorial experiment, after modular reduction to FO$^2$ with cardinality constraints and WFOMC-preserving skolemization. A clear negative correlation between them and the maximum domain size having FOMC runtime within 100 seconds (max $n$) can be observed.}
    \label{tab:active}
\end{table}

\begin{figure}[ht]
       \centering
       \subfloat{\includegraphics[width=.41\textwidth]{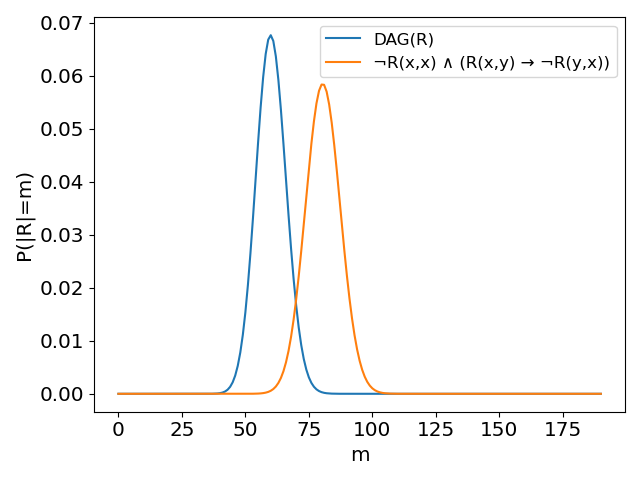}}
       \
       \subfloat{\includegraphics[width=.41\textwidth]{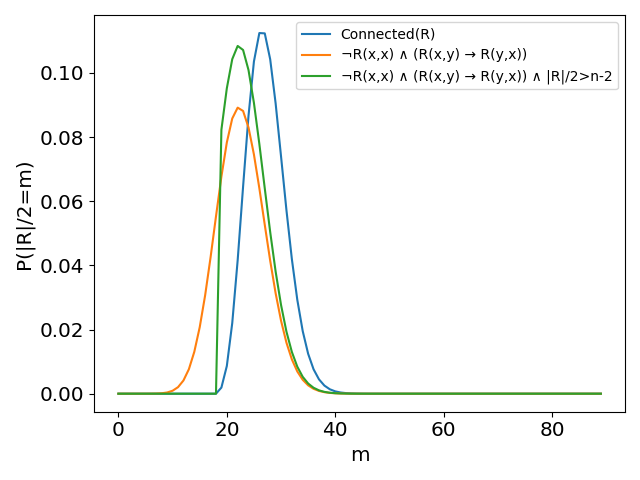}}
       \
       \subfloat{\includegraphics[width=.41\textwidth]{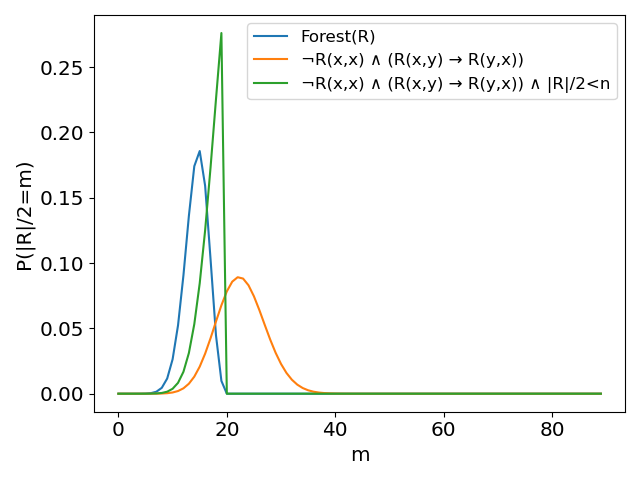}}
   \caption{Probability distributions of the number of edges of the directed/undirected graphs produced by MLNs with only the predicate $R/2$. The legends report the hard constraints, while the soft constraint is always $-1: R(x,y)$. The domain size is fixed to $n=20$.}
   \label{fig:edges}
\end{figure}

\begin{figure}[ht]
      \centering
      \subfloat{\includegraphics[width=.41\textwidth]{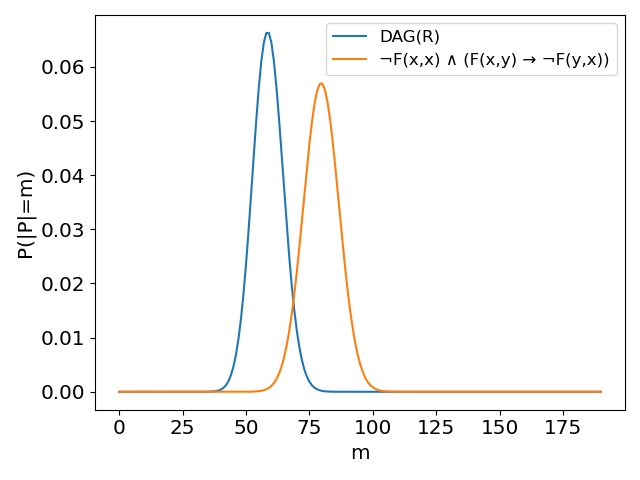}}
      \
      \subfloat{\includegraphics[width=.41\textwidth]{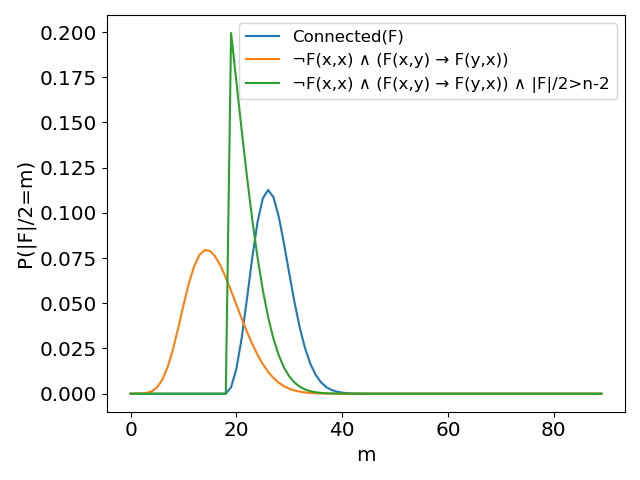}}
      \
      \subfloat{\includegraphics[width=.41\textwidth]{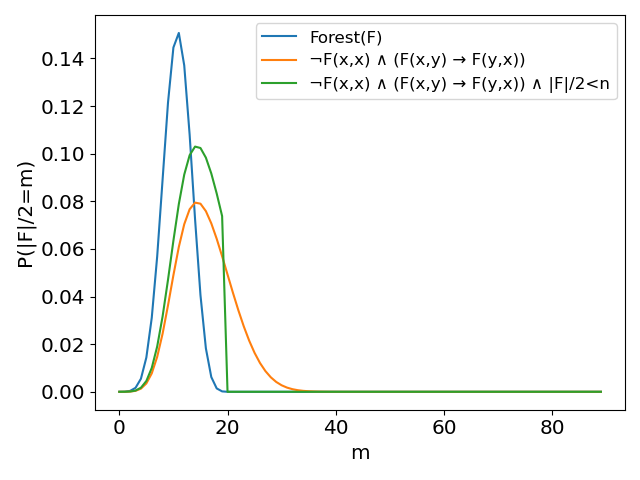}}
  \caption{Probability distributions of the number of edges of the graphs produced by predicate $F$ with ``smokers and friends'' MLNs. The legends report the hard constraints, while the soft constraints are always $0: S(x)$, $-1: F(x,y)$ and $3: P(x,y)$. The domain size is fixed to $n=20$.}
  \label{fig:edges-smokers}
\end{figure}

\begin{figure}[ht]
      \centering
      \subfloat{\includegraphics[width=.41\textwidth]{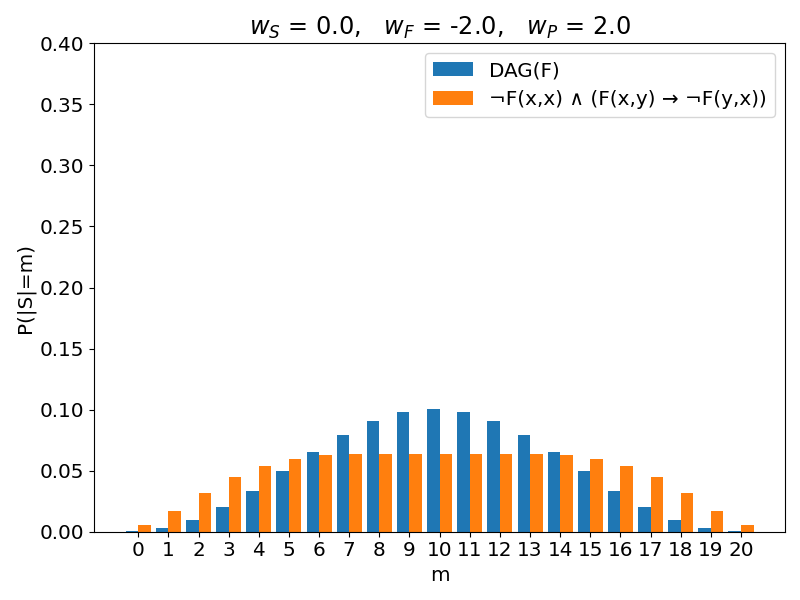}} \
      \subfloat{\includegraphics[width=.41\textwidth]{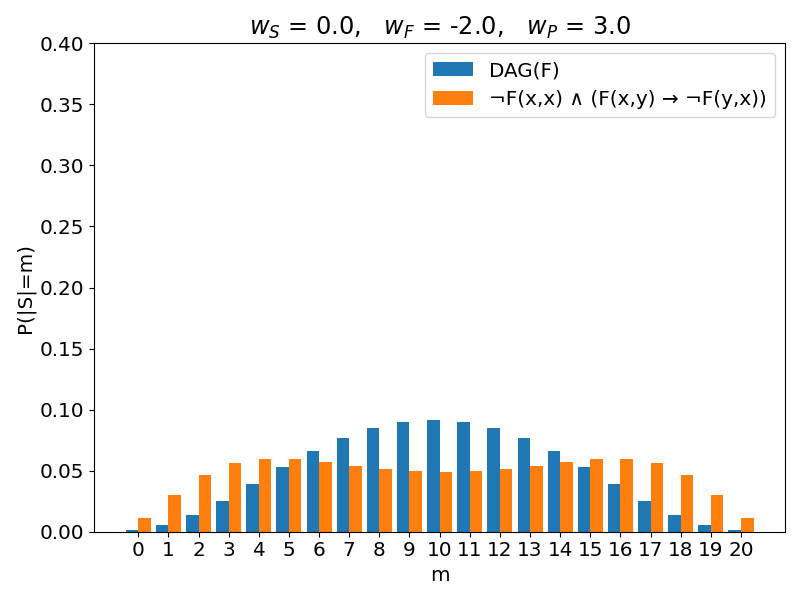}} \
      \subfloat{\includegraphics[width=.41\textwidth]{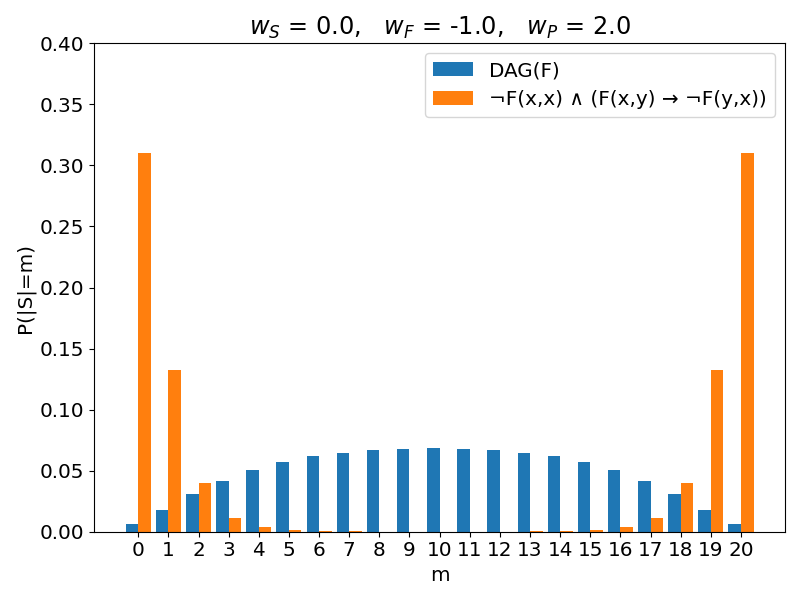}} \
      \subfloat{\includegraphics[width=.41\textwidth]{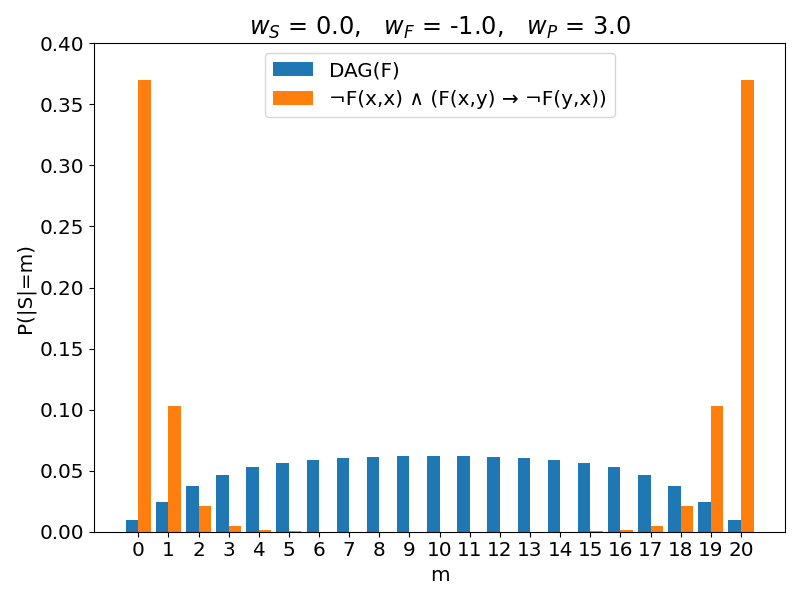}}
  \caption{Distributions of smokers for different weight values. The legends report the hard constraints, while the titles report the weights of the soft constraints. The domain size is fixed to $n=20$.}
  \label{fig:smokers-DAG}
\end{figure}

\begin{figure}[ht]
      \centering
      \subfloat{\includegraphics[width=.41\textwidth]{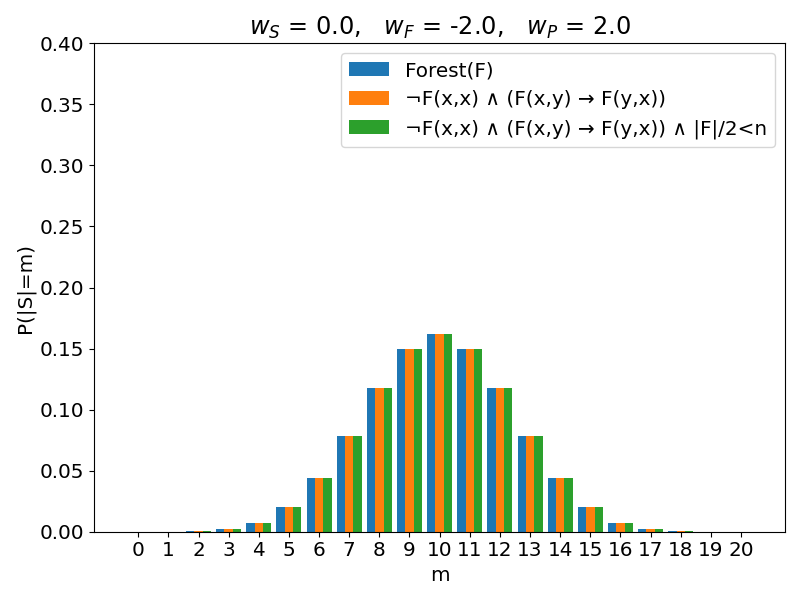}} \
      \subfloat{\includegraphics[width=.41\textwidth]{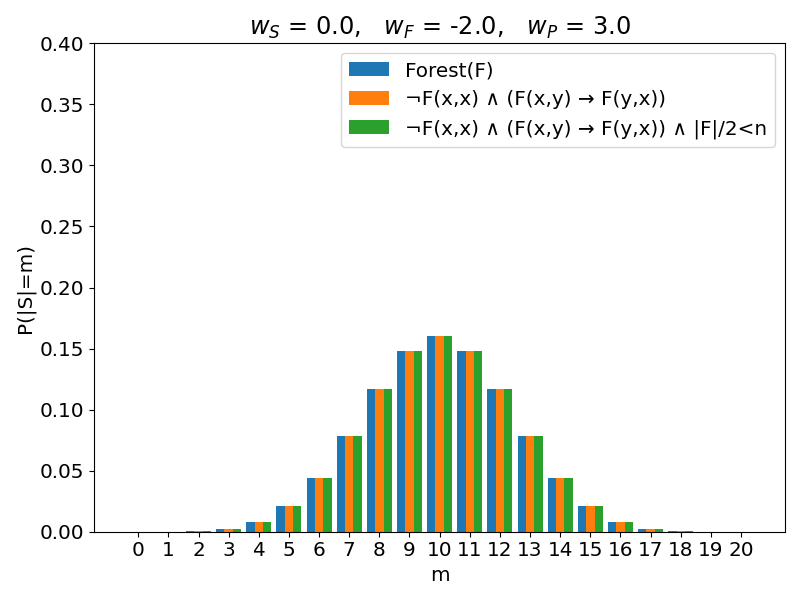}} \
      \subfloat{\includegraphics[width=.41\textwidth]{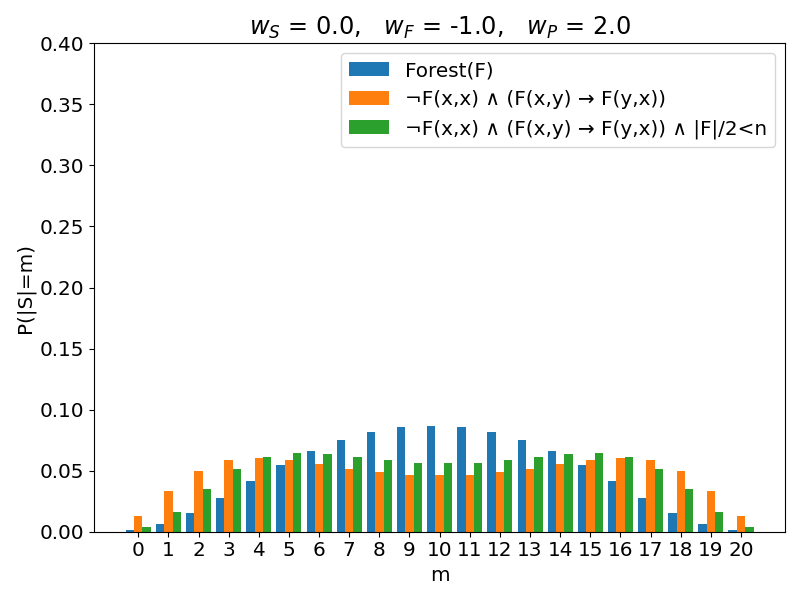}} \
      \subfloat{\includegraphics[width=.41\textwidth]{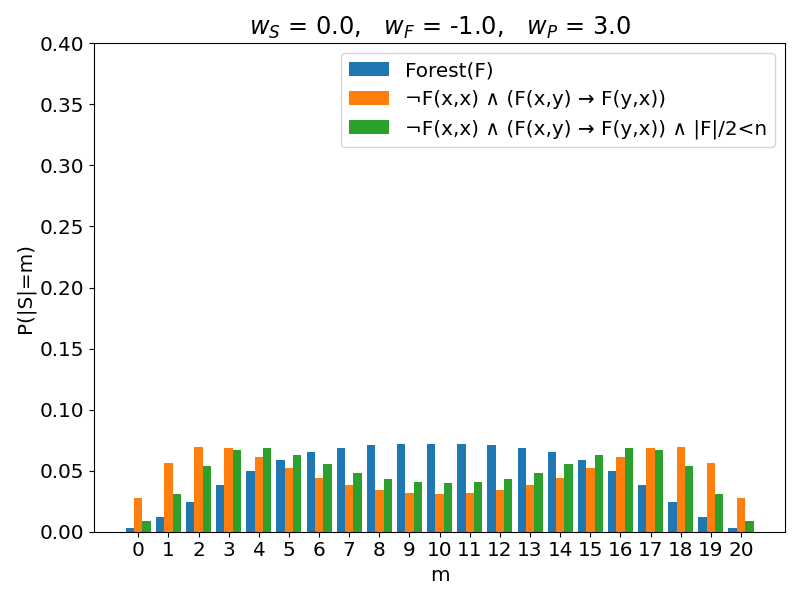}}
  \caption{Distributions of smokers for different weight values. The legends report the hard constraints, while the titles report the weights of the soft constraints. The domain size is fixed to $n=20$.}
  \label{fig:smokers-forest}
\end{figure}

\end{document}